\documentclass[journal]{IEEEtran}

\usepackage{times}
\usepackage{epsfig}
\usepackage{graphicx}
\usepackage{amsmath}
\usepackage{amssymb}
\usepackage{booktabs}
\usepackage{subfigure}
\usepackage{color,xcolor}
\usepackage[ruled,linesnumbered]{algorithm2e}
\usepackage[backref]{hyperref}

\begin{document}

% \title{Cross-Concept Feature Fusion for Generalized Zero-Shot Learning}
% \title{Evolution-guided Semantic-to-Visual Embedding Approach for Generalized Zero-Shot Learning}
% \title{Evolution-driven Multi-Knowledge Fusion for Generalized Zero-Shot Learning}
\title{Multi-Knowledge Fusion for New Feature Generation in Generalized Zero-Shot Learning}

\author{Hongxin~Xiang$^{\dag}$,
        Cheng~Xie$^{\dag,\S}$,~\IEEEmembership{Member,~IEEE,}
        Ting~Zeng,
        and~Yun~Yang$^{\S}$,~\IEEEmembership{Member,~IEEE}% <-this % stops a space
\thanks{All the authors are with National Pilot School of Software, Yunnan University, Kunming, 650504, China.}% <-this % stops a space
\thanks{$^{\dag}$Hongxin Xiang and Cheng Xie contributed equally to this work.}% <-this % stops a space
\thanks{$^{\S}$Yun Yang (Email: yangyan19@hotmail.com) and Cheng Xie (Email: xiecheng@ynu.edu.cn) are the corresponding authors.}% <-this % stops a space
% \thanks{Hongxin Xiang and Cheng Xie contributed equally to this work.}
}

\markboth{Journal of \LaTeX\ Class Files,~Vol.~14, No.~8, August~2015}%
{Shell \MakeLowercase{\textit{et al.}}: Bare Demo of IEEEtran.cls for IEEE Journals}


\maketitle

\begin{abstract}
Suffering from the semantic insufficiency and domain-shift problems, most of existing state-of-the-art methods fail to achieve satisfactory results for Zero-Shot Learning (ZSL).
In order to alleviate these problems, we propose a novel generative ZSL method to learn more generalized features from multi-knowledge with continuously generated new semantics in semantic-to-visual embedding.
In our approach, the proposed Multi-Knowledge Fusion Network (MKFNet) takes different semantic features from multi-knowledge as input, which enables more relevant semantic features to be trained for semantic-to-visual embedding, and finally generates more generalized visual features by adaptively fusing visual features from different knowledge domain.
The proposed New Feature Generator (NFG) with adaptive genetic strategy  is used to enrich semantic information on the one hand, and on the other hand it greatly improves the intersection of visual feature generated by MKFNet and unseen visual faetures.
Empirically, we show that our approach can achieve significantly better performance compared to existing state-of-the-art methods on a large number of benchmarks for several ZSL tasks, including traditional ZSL, generalized ZSL and zero-shot retrieval.
% Empirically, we show consistent improvement over existing state-of-the-art methods on the CUB and NAB datasets on the standard or generalized ZSL (GZSL) from a noisy text. In addition, we also show the advantages of the proposed method on new evaluation metrics by using five additional datasets (CUB, AwA1, AwA2, FLO and aPY), of which CUB, AwA1, AwA2 and aPY are attribute-based datasets.
\end{abstract}


\IEEEpeerreviewmaketitle

\section{Introduction}
\IEEEPARstart{W}{ith} the renaissance of deep learning, various computer vision tasks have made huge breakthroughs \cite{he2016deep, Ren2017Faster, Sun2020Few}. However, deep learning technology often relies on a large amount of labeled category data, which constitutes a serious bottleneck for building a comprehensive model for the real visual world. In order to overcome this limitation, Zero-Shot Learning (ZSL) \cite{Chao2016An,elhoseiny2017link,Zhu_2018_CVPR, xian2018zero,elhoseiny2019creativity} was proposed to build such a learning system that can predict new categories that have not been seen in the training phase. In recent years, ZSL has attracted a lot of attention due to its potential to address unseen data in the test stage.

In Zero-Shot Learning (ZSL) classification task, the model needs to identify unseen classes that never appeared in the training phase. This harsh and realistic scene is painful for traditional classification methods because there are no labeled visual data to support the training of unseen classes. In order to solve this problem, most of the existing methods use transfer learning methods, that is, assuming that the model trained on the seen class can be applied to the unseen class, and focus on learning a transferable model. The generative zero-shot learning is one of these methods.

\begin{figure}
\centering
\includegraphics[width=8.5cm,height=4cm]{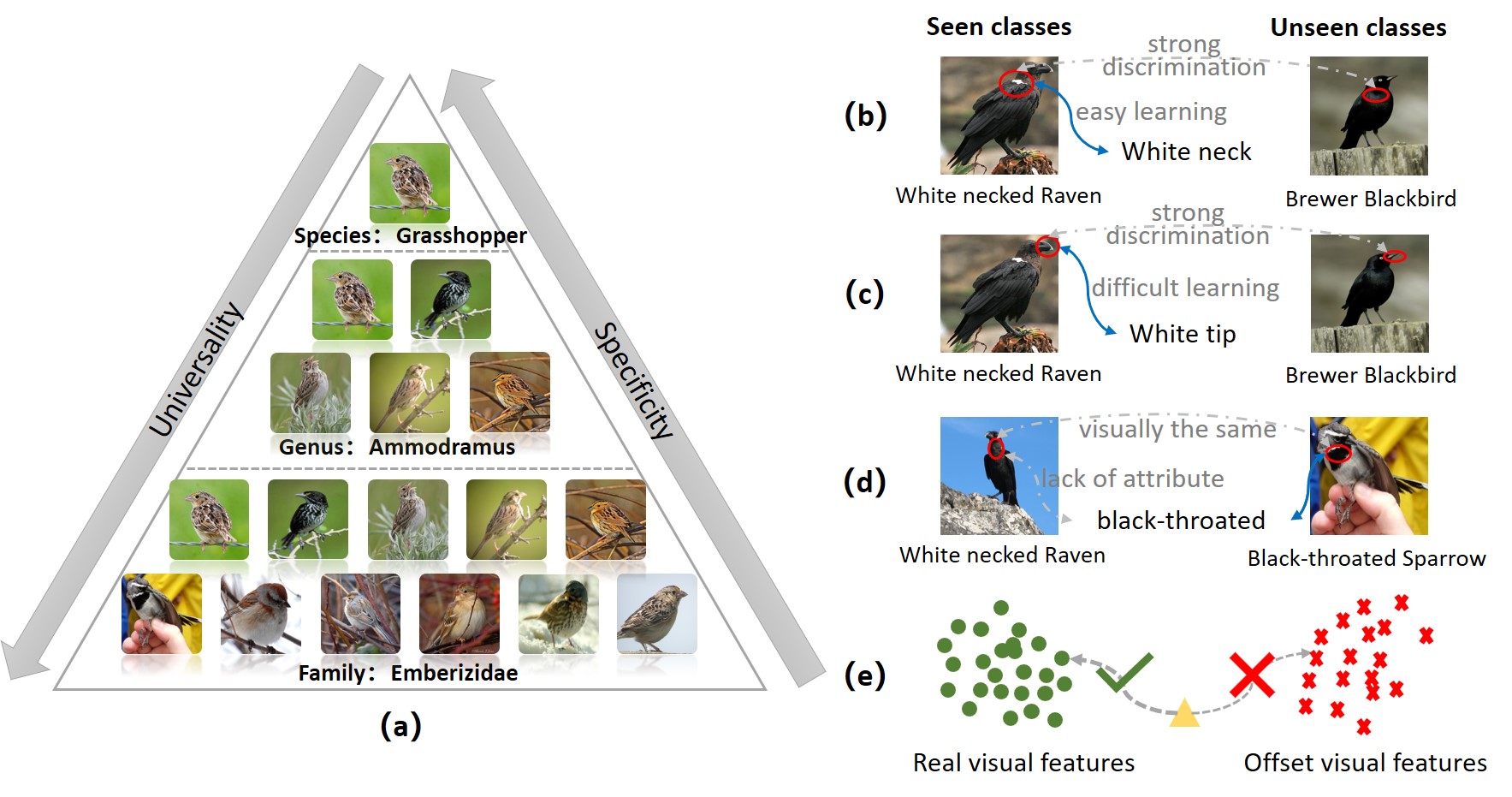}
\caption {(a) Multi-knowledge hierarchical structure of Family, Genus and Species, which is an example on the CUB dataset. The relationship of different knowledge is Family$>$Genus$>$Species, whose generality and specificity decrease and increase respectively.
(b) An example of an attribute with strong discrimination and easy learning, which shows that White neck is important attribute to distinguish White necked Raven from Brewer Blackbird.
(c) An example of an attribute with strong discrimination but difficult learning, which shows that White tip is a important attribute to distinguish White necked Raven from Brewer Blackbird but it is difficult to learn the mapping from White tip to the corresponding vision due to visually not obvious.
(d) An example of exists visually but not semantically, which shows that the attribute black-throated exists visually in both the seen class White necked Raven and the unseen class Black-throated Sparrow, but the semantics does not exist in the seen class.
(e) An example of limiting the generation of visual samples, which shows that when the generated samples are close to the corresponding clusters, they should deviate from those offset visual samples.
}
\label{Intro}
\end{figure}

The goal of generative zero-shot learning is to establish a semantic-to-visual mapping, and train a classification model that recognizes unseen classes by using synthesized-visual features transformed from unseen classes' semantics. However, for almost all semantic-to-visual feature learning methods, they still face two major challenges:

\emph{1) The Semantic Insufficiency Challenge:} We often use semantic-visual embedding, including embedding from semantic space to visual space \cite{weston2010large,akata2013label,Zhong2020Multi,min2020domain} or vice versa \cite{Zhu_2018_CVPR,elhoseiny2019creativity,zhu2019generalized,Guo2020Attentive,huynh2020fine} or to a shared common space \cite{liu2020hyperbolic,akata2015evaluation,fu2015zero,qi2016joint}, to extract the relationship between visual features and semantic features. However, semantic-visual embedding always faces the challenge of semantic insufficiency, which will lead to incomplete expression of semantic and visual features in embedding. Especially for two similar categories, it is difficult to learn discriminative features. To solve this problem, many effective methods have been proposed to improve the expression of semantic and visual features, including attention-based methods \cite{zhu2019semantic,ji2018stacked,huynh2020fine,liu2019attribute} and knowledge graph-based methods \cite{kampffmeyer2019rethinking,lee2018multi,liu2020attribute}. Attention-based methods enhance the expression of visual and semantic features by making the model focus on more fine-grained visual or semantic features; Knowledge graph-based methods enrich the expression of semantic and visual features by aggregating node information around categories. However, the semantic diversity of these two methods is limited by the dataset itself and does not generate any new semantic information. We summarized several phenomena that need to be improved in semantic-to-visual embedding: 1) The model needs to map more discriminative semantic features to visual features. As shown in Fig.\ref{Intro}(b), "White nick" is an important attribute that distinguishes "White necked Raven" from "Brewer Blackbird". The proposed method enhances the learning of this attribute by increasing the importance of this attribute and reducing the importance of others; 2) The model needs to have a stronger ability to find insignificant semantic-to-visual embedding. As shown in Fig.\ref{Intro}(c), the "White tip" attribute is difficult to correspond visually because it is not obvious visually. The proposed method implicitly reduces the importance of other attributes to force the model to learn its semantic-to-visual mapping relationship.
% 我们总结了语义嵌入过程中几种仍待改进的现象：1) 模型需要将更多的有判别性的语义特征映射到视觉特征。如图2(b)显示，white nick是White necked Raven区别于Common Raven的重要属性。提出的方法通过提高该属性的重要性或减少其它属性的重要性是有必要的对于增强该属性的学习；2) 模型需要有更强的找出不显著语义嵌入的能力。如图2(c)显示，White tip属性在视觉上很难对应由于视觉上地不明显。提出的方法隐含地降低其它属性的重要性来迫使模型学习其语义到视觉映射的关系。

\emph{2) The Domain-Shift Challenge:} The domain-shift challenge in ZSL was first identified by \cite{fu2015transductive}, which is caused by the huge difference between the seen and unseen classes in the semantic space and visual space. For example, the "tails" of pigs and zebras are not semantically different, but they are very different visually. To solve this problem, many transductive ZSL approaches \cite{7053935,FuTransductive,guo2016transductive,YuTransductive} have been proposed. They integrate the data of seen and unseen categories to learn more general visual-semantic embeddings. However, this method is unreasonable because it is difficult for us to get the relevant data of unseen classes from the beginning in the real scene. Therefore, in most cases, these methods do not have strong generalization. Similarly, we also summarized several unresolved phenomena in the domain-shift: 1) The seen classes have the visual features of the unseen classes, but it lacks the semantic information of the unseen classes in semantic-to-visual embedding. As shown in Fig.\ref{Intro}(d), the attribute "black throated" does not exist in the seen class "White necked Raven" but exists in the unseen class "black-throated sparrow", they all have the same visual features. Our method alleviates this phenomenon by generating new and reliable semantic features; 2) The unrestricted search space of the model leads to more prone to domain-drift. As shown in Fig.\ref{Intro}(e), our approach limits the visual space by generating offset visual features, so that the synthesized visual features can deviate to the unseen category.
% 同样，我们也总结了领域飘逸问题中几个有待解决的现象：1) 可见类中存在未见类的视觉特征而缺少未见类的语义描述在语义到视觉嵌入时。如图2(d)所示，属性"black throated"在可见类"White necked Raven"中不存在而在未见类"black-throated sparrow"中存在，他们都有同样的视觉特征。我们的方法通过生成新的可靠的语义特征来缓解这一现象。；2) 模型的搜索空间不受限导致更容易发生领域飘逸现象。如图2(e)所示，我们的方法通过生成偏移的视觉特征来限制视觉空间，使得生成的视觉向量向未见类偏移。

In this paper, in order to alleviate these two challenges, we propose a new generative zero-shot learning method. The proposed method uses the two modules, including Multi-Knowledge Fusion Network (MKFNet) and New Feature Generator (NFG), to deal with the semantic insufficiency challenge and domain-shift challenge.
Following our previous work \cite{xie2021cross}, the proposed method first builds a three-level knowledge hierarchy according to the taxonomic Family-Genus-Species knowledge. Note that the level from low to high is Species, Genus, Family. Compared with the low-level knowledge, the high-level knowledge have higher universality or generalization and lower specificity for model learning, as shown in Fig.\ref{Intro}(a). Then the MKFNet is trained by using different semantic features from multi-knowledge domain to obtain more semantic information, and the proposed knowledge regularization term $L_{KR}$ is used to generate visual features belonging to different knowledge, which will imply unseen information. Finally, more generalized visual features are generated by adaptively fusing different synthesized visual features from multi-knowledge. Meanwhile, the proposed NFG adds new semantic features including enhanced and novel ones. The enhanced regularization term $L_{ER}$ and novel regularization term $L_{NR}$ are used to further enrich semantic knowledge and generate visual features that deviate to unseen samples. In summary, the main contributions of this paper are as follows:

\begin{itemize}
  \item We propose a new generative ZSL model, Multi-Knowledge Fusion Network (MKFNet), to adaptively fuse different visual features from multi-knowledge domain for synthesizing more generalized visual features.
  % We propose a new generative ZSL model, Multi-Knowledge Fusion Network (MKFNet), to learn more generalized visual features, which are synthesized by adaptive fusion module to fuse different visual features from multi-knowledge domain. These visual features from different knowledge are generated by knowledge regularization term $L_{KR}$. % They all obviously enrich the semantic features to deal with the semantic insufficiency challenge.

  \item We propose the New Feature Generator (NFG) to generate new semantic features for each knowledge domain, including enhanced semantic features and novel semantic features, to make the MKFNet learn more semantic information and generate visual features that intersect with unseen classes.
  % We propose the New Feature Generator (NFG) with enhanced regularization term $L_{ER}$ and novel regularization term $L_{NR}$ to generate more semantic features, including enhanced semantic features and novel semantic features, by using adaptive genetic algorithm. The $L_{ER}$ and $L_{NR}$ are respectively used to supervise enhanced and novel semantic features to obtain more semantic information and generate visual features that intersect with unseen classes.

  \item Extensive experiments on seven major benchmarks show that the proposed approach achieves state-of-the-art performances on both the traditional ZSL and the generalized ZSL tasks.
\end{itemize}

% 2. 相关工作
\section{Related Works}

% 生成式
\subsection{Generative ZSL}
Recently, generative methods dominate ZSL, which reduces ZSL to a traditional classification problem. \cite{long2017zero,Zhu_2018_CVPR,xian2018feature,kumar2018generalized,han2020learning} used the existing generative model or its variants to synthesize visual features by using category semantics and random noise. The difference is that \cite{long2017zero} proposed a one-to-one mapping approach where synthesized visual features are restricted; \cite{Zhu_2018_CVPR} introduced visual center regularization to make the synthesized visual features close to the center of the class; \cite{xian2018feature} improved the performance of the generator by adding auxiliary classifier; \cite{kumar2018generalized} used the encoder-decoder structure to constrain the generation of visual features, so as to get a better generator; \cite{han2020learning} projected the original visual features into a new (redundancy-free) space to learn the redundancy-free features. Our model is also a generative method, which generates more generalized visual features by using semantic features from multi-knowledge as input. In the previous generative methods, several are closely related to our model. For example, GDAN \cite{Huang_2019_CVPR} used semantic-to-visual and visual-to-semantic multiple mapping to retrain enough semantic information; DGP \cite{kampffmeyer2019rethinking} and APNet \cite{liu2020attribute} obtained more semantic features by constructing a knowledge graph and aggregating the information of neighbor nodes. Although this method can alleviate the problem of semantic insufficiency and domain shift to a certain extent, it does not directly produce informations of unseen classes. In contrast, our method compensates for the lack of semantics by adding semantic information of different knowledge and adaptively generating more semantic features. In addition, the proposed approach uses $L_{KR}$ regularization to enable our synthesized samples to deviate to unseen classes.

\subsection{Attention-based approach}
The purpose of the attention mechanism is to highlight important local information, or to shield the influence of irrelevant information or noise information. Due to its effectiveness and versatility, the attention mechanism has been widely used in various computer vision tasks, such as image classification\cite{wang2017residual,guo2019visual}, image segmentation\cite{chen2016attention,nie2018asdnet}, image generation\cite{xu2018attngan, gao2020lightweight}, etc. Recently, a large number of attention-based methods \cite{zhu2019semantic,ji2018stacked,huynh2020fine,liu2019attribute} have been proposed for ZSL recognition tasks. For instances, \cite{zhu2019semantic} proposed a stacked semantics-guided attention (S$^2$GA) model to obtain semantic related features by using individual category semantic features to progressively guide visual features to generate attention maps for weighting the importance of different local regions; \cite{huynh2020fine} proposed a dense attribute-based attention mechanism, which focuses on the most relevant image regions for each attribute and obtains attribute-based features. Although these attention-based methods improved the performance of the model by mining local semantic or visual features, they easily lead to overfitting in semantic-visual embedding. In this paper, we propose a Multi-Knowledge Fusion Network (MKFNet) with adaptive fusion module based on the attention mechanism, which combines the visual features of different knowledge and tries to find the optimal visual weight between different knowledge to generate more generalized samples.

\subsection{Genetic-based Approach}
Genetic algorithm is an randomized and optimized method guided by the principles of evolution and natural genetics. It is widely used in various tasks \cite{Sun2020Automatically,Ma2020Autonomous,Li2018Genetic,Ye2019An} because of its high efficiency, adaptability and robustness. And it can produce near-optimal solutions and can handle large, highly complex and multi-modal spaces. For instances, \cite{Sun2020Automatically} proposed an automatic CNN architecture design method by using genetic algorithms, to effectively address the image classification tasks. \cite{Ma2020Autonomous} used genetic evolutionary operations, including selection, mutation and crossover to evolve a population of DCNN architectures.
To our best knowledge, there is no work to utilize the genetic algorithm to ZSL task. In our work, we proposed a New Feature Generator (NFG) with adaptive genetic algorithm for ZSL, which uses mutation and crossover strategies to generate new semantic features and uses a similarity-based selection strategy to adaptively filter new semantic features into enhanced and novel semantic features. Specially, NFG uses two regularization terms $L_{ER}$ and $L_{NR}$ to further alleviate the problems of semantic insufficiency and domain-shift, respectively.

% 3. 方法
\section{The Proposed Approach}

In this section, we first introduce the overall architecture of the proposed approach in Section \ref{ModelArchitecture}. Then the Multi-Knowledge Fusion Network (MKFNet) and the implementation of the adaptive fusion module are explained in Section \ref{CCFFNet} and Section \ref{AFM}, respectively. After that, New Feature Generator (NFG) is described in detail in Section \ref{AGS}. Finally, we discuss the training and testing process of the proposed approach in Section \ref{TrainTest} .

\begin{figure*}
\centering
\includegraphics[width=18cm,height=8cm]{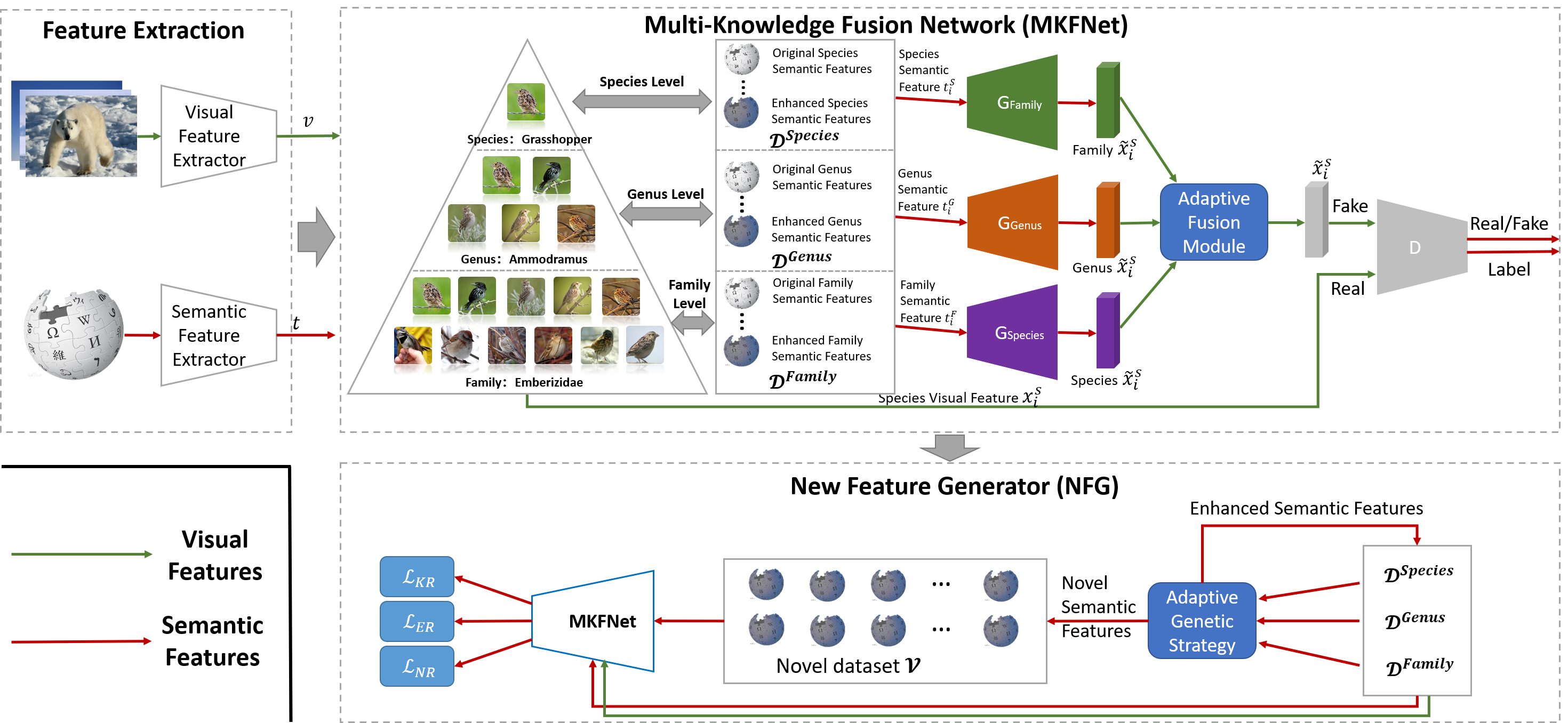}
\caption {The overall framework of the proposed approach.}
\label{modelFrame}
\end{figure*}

%\subsection{Problem Formulation}
%Let $\mathcal{T}$, $\mathcal{X}$ and $\mathcal{Y}$ represent the semantic space, visual space and label space, respectively. $t_i^u\in \mathcal{T}^u$ and $t_i^s\in \mathcal{T}^s$ represent the i-th semantic feature of unseen classes and seen classes, respectively. $x_i^s\in \mathcal{X}^s$ and $x_i^u\in \mathcal{X}^u$ represent the i-th visual feature of seen classes and unseen classes, respectively. $y_i^s\in \mathcal{Y}^s$ and $y_i^u\in \mathcal{Y}^u$ represent the i-th label of unseen classes and seen classes, respectively. Note that the seen and the unseen classes are disjointed, i.e., $\mathcal{Y}^s \bigcap \mathcal{Y}^u=\varnothing$. All sample points $(x_i, y_i, t_i)$ are sampled from whole dataset $\mathcal{D}=\mathcal{D}^s\cup \mathcal{D}^u$. The normal population $\mathcal{V}$ and novel population $\mathcal{W}$ represent the dataset used by AGS. They are initialized to the original seen dataset $\mathcal{D}^s$ and the empty set $\phi$ respectively at the beginning. The zero-shot learning (ZSL) and generalized zero-shot learning (GZSL) classification tasks are to given the unseen classes visual samples $x^u$ to predict the labels $y\in \mathcal{Y}^u$ and the labels $y\in \mathcal{Y}^u\cup \mathcal{Y}^s$, respectively. Obviously the GZSL classification task is more challenging.
\subsection{The Model Architecture}
\label{ModelArchitecture}
The Fig.\ref{modelFrame} shows the model architecture of the proposed method. It is obviously that the proposed approach is based on generative model and our Multi-Knowledge Fusion Network (MKFNet) consists of three generators belonging to different knowledge and one discriminator. The generator of the GAZSL \cite{Zhu_2018_CVPR} is selected as skeletons of generators $G_{Family}, G_{Genus}$ and $G_{Species}$, and the discriminator of GAZSL is selected as skeleton of the discriminator of the proposed MKFNet. The detailed process of the proposed method is described as the following steps:

\textbf{Feature Extraction:} For any dataset, we need to extract visual features from images and semantic features from attributes or text, respectively. For details, see Semantic Representation and Visual Representation in \ref{ExperimentalSettings}.
% For CUB and NAB datasets, we use the part-based FC layer in VPDE-net \cite{zhang2017learning} as the visual feature extractor to obtain fine-grained visual features and use Term Frequency Inverse Document Frequency (TF-IDF) \cite{TF-IDF} as the semantic feature extractor to obtain semantic features according to the related work \cite{Elhoseiny2013Write, ElhoseinyLink}. For CUB-Att, AwA1, AwA2, FLO and aPY, the 2,048-dimensional visual features are extracted by ResNet-101 \cite{he2016deep} pretrained on ImageNet. The semantic features are extracted by pre-defined attributes on each dataset.
% 对于任意数据集，我们将提取视觉特征从图像中和语义特征从属性或文本中，详细见\ref{ExperimentalSettings}.

\begin{figure}
\centering
\includegraphics[width=6cm,height=6cm]{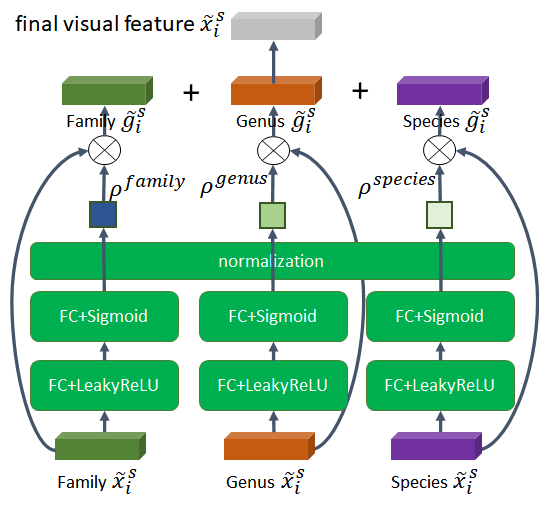}
\caption {The detail of the adaptive fusion module.}
\label{FusionModel}
\end{figure}

\textbf{Multi-Knowledge Fusion Network (MKFNet):} Following our previous work \cite{xie2021cross}, the hierarchy of the three knowledge (Species, Genus, Family) is first constructed according to class names and biological taxonomy. Then, three semantic features of different knowledge are input into the corresponding generators to synthesize visual features $\tilde{x}_i^{Species}, \tilde{x}_i^{Genus}$ and $\tilde{x}_i^{Family}$, and the adaptive fusion module will weight these synthesized visual features to obtain the final visual features $\tilde{x}_i$, as shown in the Fig.\ref{FusionModel}. The real visual features of the corresponding species and the final visual features are input into the discriminator for real and fake discrimination and label classification.

\textbf{New Feature Generator (NFG):} NFG uses the original semantic features from different knowledge as the input of the adaptive genetic strategy (see Fig.\ref{AGSFig} for details) to generate more new semantic features, including enhanced semantic features and novel semantic features. Enhanced semantic features are used to supplement more semantic information for each original knowledge domain, and new semantic features are used for unsupervised training of MKFNet to improve the generalization ability of the model.
% For each category of different knowledge (Species, Genus, Family), NFG is used to generate more enhanced semantic features and novel semantic features by using adaptive genetic algorithm. The enhanced semantic features are added to original semantic features and the novel semantic features are used as input to the discriminator for unsupervised training.

% \textbf{Cross-Concept Training Scheme(CCTS):} For a given training data point, CCTS enriches the semantic feature for each visual feature by extracting extra text or attributes from cross concepts.

\subsection{Multi-Knowledge Fusion Network}
\label{CCFFNet}
In order to solve the zero-shot classification task, a model needs to be trained to infer the unseen categories from the corresponding class semantic prototypes. To achieve this goal, we design a Multi-Knowledge Fusion Network (MKFNet) to map semantic features into visual space. In short, MKFNet is a generative adversarial network, which consists of three generators and a discriminator.

Previous zero-shot learning methods usually use a single semantic feature to learn semantic-to-visual embedding, which obviously leads to insufficient semantics. We follow our previous work \cite{xie2021cross} to use a variety of relevant semantic features to learn, which obviously increase more semantic information. Following \cite{xie2021cross}, we can construct a Family-Genus-Species hierarchy of three knowledge for original dataset according to taxonomy. The hierarchical relationship between different knowledge is Family $>$ Genus $>$ Species. Higher level knowledge include lower level knowledge, such as Genus are subsets of Family and Species are subsets of Genus. The final structure is shown in Fig.\ref{Intro}(a). We formalize the datasets $\mathcal{D}^{Family}, \mathcal{D}^{Genus}, $ and $\mathcal{D}^{Species}$ of different knowledge as:

\begin{equation}
\label{DS}
   \mathcal{D}^{\bullet}=(x_i^s, y_i^s, t_i^{s,\bullet}), i=1,...,N^s
\end{equation}
\iffalse
\begin{equation}
\label{DG}
   \mathcal{D}^{Genus}=(x_i^s, y_i^s, t^{s,Genus}), i=1,...,N^s
\end{equation}

\begin{equation}
\label{DF}
   \mathcal{D}^{Family}=(x_i^s, y_i^s, t^{s,Family}), i=1,...,N^s
\end{equation}
\fi
Where $N^s$ is the total number of seen class samples, $\bullet$ can be represented Species, Genus and Family. $t^{s,Species}$, $t^{s,Genus}$ and $t^{s,Family}$ denote semantic features belonging to the same Species, Genus and Family, respectively. Note that the $y_i^s$ in Genus and Family will be replaced with the new class labels of corresponding Genus and Family. Obviously, visual feature $x_i^s$ in the same Family will include visual features of multiple Genus or Species.

MKFNet uses three generators to learn information about three different knowledge, which are $G_{Species}, G_{Genus}$ and $G_{Family}$. The generator of each knowledge is used to map semantic features to visual space $\mathbb{R} ^S\rightarrow\mathbb{R}^V$. Obviously, generator with higher knowledge will learn more generalized visual features. In order to make each model learn the semantic-to-visual mapping of different knowledge better, the knowledge regularization term is added to these generators to make the synthesized visual features close to the visual center of the corresponding knowledge. The knowledge regularization terms of generators $G_{Species}, G_{Genus}$ and $G_{Family}$ are formalized as follows:

\begin{equation}
\label{LCRS}
   L_{KR}^{Species}=\frac{1}{N^s}\sum_i^{N^s} \|G_{Species}(t_i^s, z)-\hat{x}_i^{s,Species}\|^2
\end{equation}
\begin{equation}
\label{LCRG}
   L_{KR}^{Genus}=\frac{1}{N^s}\sum_i^{N^s} \|G_{Genus}(t_i^s, z)-\hat{x}_i^{s,Genus}\|^2
\end{equation}
\begin{equation}
\label{LCRF}
   L_{KR}^{Family}=\frac{1}{N^s}\sum_i^{N^s} \|G_{Family}(t_i^s, z)-\hat{x}_i^{s,Family}\|^2
\end{equation}
Where $N^s$ is the total number of seen samples, and $G_\bullet(t_i,z)$ is the visual feature generated by $G_\bullet$ according to semantic feature $t_i$ and random noise $z$. $\hat{x}_i^{s,Species}$ represents the visual feature center of all categories belonging to $y_i^s$'s Species. Similarly, $\hat{x}_i^{s,Genus}$ and $\hat{x}_i^{s,Family}$ denote the visual center of Genus and Family corresponding to $y_i^s$, respectively. $L_{KR}$ regularization force the synthesized visual features to deviate to the unseen class by closing to the class center of higher level knowledge, which increases the generalization ability of the generator. The loss of these generators are defined as:
\begin{equation}
\label{GLossSpecies}
   L_{G}^{\bullet}=-\mathbb{E}[D(G_{\bullet}(\mathcal{T}, z))]+L_{cls}(G_{\bullet}(\mathcal{T}, z))+L_{KR}^{\bullet}
\end{equation}
Where the first term is Wasserstein loss \cite{Gulrajani2017Improved}, and the second term is cross-entropy loss of classification. $\bullet$ can be represented as Species, Genus, or Family. For example, $L_{G}^{Species}$ represents the loss of $G_{Species}$.

After generating three visual features $\tilde{x_i}^{s,Species}, \tilde{x_i}^{s,Genus}$ and $\tilde{x_i}^{s,Family}$ belonging to different knowledge by using $G_{Species},G_{Genus}$ and $G_{Family}$, the final synthesized visual feature $\tilde{x_i^s}$ is obtained through an attention-based adaptive fusion module, which fuses visual information of different knowledge. See section \ref{AFM} for details. $D$ is the discriminator, which has the same structure as the discriminator of ACGAN \cite{ShlensConditional}. The discriminator $D$ first receives the final synthesized visual features $\tilde{x_i^s}$ and real visual features corresponding to Species as input. Then these two visual features are forwarded through two full connected layers. Subsequently, there are two output branches: the first branch is used for binary classifier to distinguish whether the input features are real or fake; the second one classifies the input samples into correct classes. The loss function of the discriminator $D$ is the same with the previous work \cite{Zhu_2018_CVPR}.

\subsection{Adaptive Fusion Module}
\label{AFM}
To solve the problem of feature fusion from different knowledge, features summing is a simple and common approach, which can perform the summing operation on the output $\tilde{x_i}^{s,Species}$, $\tilde{x_i}^{s,Genus}$ and $\tilde{x_i}^{s,Family}$ from $G_{Species}, G_{Genus},$ and $G_{Family}$ to obtain the fusion feature $\tilde{x_i^s}$. We can easily perform this operation because the output of each generator has the same dimensions as the visual features. The operation is formalized as:
\begin{equation}
\label{fusionFormla}
   \tilde{x_i^s}=\frac{\tilde{x_i}^{s,Family}+\tilde{x_i}^{s,Genus}+\tilde{x_i}^{s,Species})}{3}
\end{equation}
% Here, we present our baseline fusion model via features summing, which is a straightforward approach to perform the summing operation on the output of $G_{Species}, G_{Genus},$ and $G_{Family}$. We can easily perform this operation because their output is all the same dimension as the visual features.

However, the features summing method is unreasonable that the above-mentioned fusion method treat three visual features from different knowledge with equal importance. In practice, due to the inconsistent distribution of different knowledge domains, these visual features from different knowledge has different contributions to the final synthesized visual features. For example, the universality of Family, Genus, and Species sequentially decreases, and the specificity increases sequentially. Therefore, it is very necessary to assign weights reasonably to visual features of different knowledge. To handle these situation, we propose an adaptive fusion module to automatically assign importance weights to these three visual features. Inspired by attention mechanism \cite{wang2017residual,guo2019visual}, we apply the same ideas into our method to learn the importance weights. Fig.\ref{fusionFormla} visually shows the overall architecture of the adaptive fusion module. The input features $\tilde{x_i}^{s,Species}, \tilde{x_i}^{s,Genus}$ and $\tilde{x_i}^{s,Family}$ are projected by a full connected layer with LeakyReLU into three vectors of the same dimensionality. Let \textbf{$W^{\bullet}_1$} and \textbf{$b^{\bullet}_1$} denote the projection (weight) matrix and the bias vector of the first full connected layer that directly takes the visual features of Family, Genus and Species as the input. $\bullet$ can be represented as Family, Genus and Species. Subsequently, the three projected visual features are passed through a nonlinear activation function $\sigma_1$, which is chosen to be the LeakyReLU function in this paper. The output of the first layer is formalized as \textbf{$\sigma_1(W^{\bullet}\tilde{x_i}^{s,\bullet}+b^{\bullet})$}.

Furthermore, the output of the first layer is propagated forward to a full connected layer, which has only one output neuron. We define the weight matrix and bias vector of the second layer as \textbf{$W^{\bullet}_2$} and \textbf{$b^{\bullet}_2$}, and the activation function as $\sigma_2$ (which is Sigmoid in this paper). As a result, the importance weight is predicted as a scalar score, which is formalized as:
\begin{equation}
   s^{\bullet}=\sigma_2(W_2^{\bullet}(\sigma_1(W_1^{\bullet}\tilde{x_i}^{s,\bullet}+b_1^{\bullet}))+b_2^{\bullet})
\end{equation}

In order to ensure the consistency of the visual features before and after the fusion, the normalization layer is used to normalize the three scalars. The normalized scalar is used as the final importance weight, denoted by $\rho^{Family}$, $\rho^{Genus}$, and $\rho^{Species}$ respectively. They are formalized as:
\begin{equation}
   \rho^{\bullet}=\frac{s^{\bullet}}{s^{Family}+s^{Genus}+s^{Species}}
\end{equation}

Finally, we can obtain the fusion feature $\tilde{x_i^s}$ by a weighted sum, which is formulated as:

\begin{equation}
\label{fusionFormla}
   \tilde{x_i^s}=\rho^{Family}\tilde{x_i}^{s,Family}+\rho^{Genus}\tilde{x_i}^{s,Genus}+\rho^{Species}\tilde{x_i}^{s,Species}
\end{equation}

The loss function of the adaptive fusion module is formalized as:

\begin{equation}
\label{LFM}
   % L_{FM}=-\mathbb{E}[D(\tilde{x^s})]+L_{cls}(\tilde{x^s})
   L_{FM}=-\mathbb{E}[D(\tilde{x^s})]-\mathbb{E}[\Sigma_{k=1}^{K^s}y_k \log(D(\tilde{x^s}))]+L_{ER}+L_{NR}
\end{equation}
Where, the first term tricks the discriminator to classify synthesized visual features from generator as real and the second term is classified cross-entropy loss. The third and forth terms will describe in Section \ref{AGS}. $\tilde{x^s}$ is obtained according to the Formula \ref{fusionFormla}.

% \textbf{Feature Weight Prediction:}
% \textbf{Feature Weighted Fusion:}

% 描述跨多个level的网络，使用了注意力机制。

\subsection{New Feature Generator}
\label{AGS}

\begin{figure*}
\centering
\includegraphics[width=14cm,height=6cm]{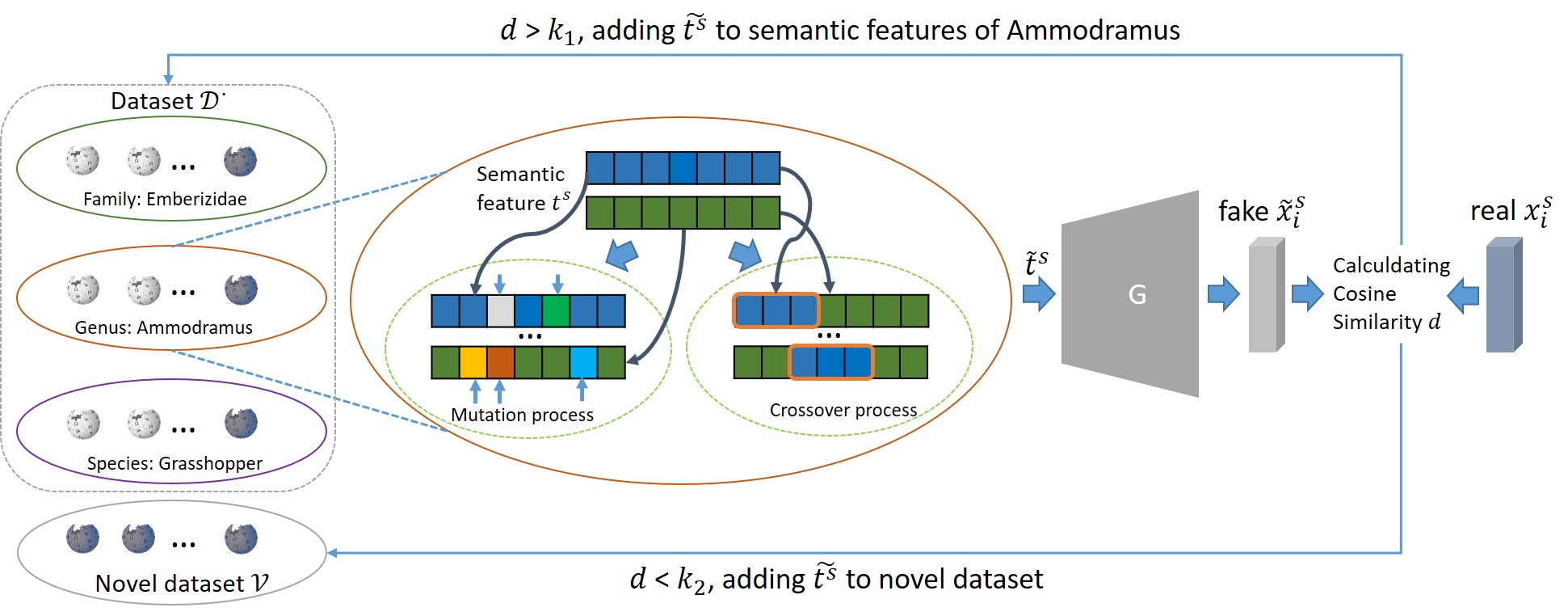}
\caption {The whole process of the adaptive genetic strategy. Take Genus Ammodramus as an example. First, randomly sample two semantic features $t^s_a$ and $t^s_b$ from the semantic features of Ammodramus, and use them to generate new semantic features $\tilde{t^s}$ through the mutation process and the crossover process, see Algorithm \ref{AGSAlgo} for details. Then, the cosine similarity $d$ is calculated by using the synthesized visual feature, which is generated by inputting the new semantic feature $\tilde{t}^s$ into MKFNet, and the real visual feature. Finally, if the $d$ is greater than the stability coefficient $\kappa_1$, it is regarded as enhanced semantic features and added to the semantic features of Ammodramus. If the $d$ is less than the stability coefficient $\kappa_2$, it is regarded as novel semantic features and added to the novel dataset. $\kappa_1$ and $\kappa_2$ satisfy $\kappa_1>\kappa_2$.}
% 以Genus为Ammodramus为例。首先，从Genus为Ammodramus的语义特征中随机采样2个语义特征$t^s_a$和$t^s_b$，并使用他们去生成新的语义特征$\tilde{t^s}$通过突变过程和交叉过程，详细见算法5。然后，计算余弦相似度$d$通过使用生成的视觉特征，它被生成通过将新的语义特征$\tilde{t}^s$输入到MKFNet的生成器中，和真实的视觉特征。最终，如果余弦相似度$d$大于稳定系数$\kappa_1$，则被视为enhanced semantic features加入到Ammodramus的语义特征中。如果余弦相似度$d$小于稳定系数$\kappa_2$，则被视为novel semantic features加入到Ammodramus的语义特征中。$\kappa_1$和$\kappa_2$满足$\kappa_1>\kappa_2$。
\label{AGSFig}
\end{figure*}

The whole process of New Feature Generator (NFG) is shown in Fig.\ref{AGSFig}.
% For ease of understanding, we first divide the dataset into two major parts, the enhanced dataset $\mathcal{V}$ initialized by the original seen class samples $D^{Species}, D^{Genus}$ and $D^{Family}$, and the novel population $\mathcal{W}$ that has shifted to the unseen class due to abnormal inheritance.
In order to obtain the new semantic features $\tilde{t}^s$ from different knowledge, we first randomly sample two semantic features $t_a^s, t_b^s\in \mathcal{T}$ from any category of $\mathcal{D}^{Species}, \mathcal{D}^{Genus}$ and $\mathcal{D}^{Family}$. Note that every dataset $\mathcal{D}^{\bullet}$ should be considered. In particular, the Fig.\ref{AGSFig} shows the new semantic generation process with the Genus label Ammodramus, so two semantic features need to be randomly sampled from all the semantic features with the Genus label Ammodramus. Then, the semantic feature $\tilde{t}^s$ is obtained through two operations: the mutation of a single semantic feature $t_a^s$ or $t_b^s$ and the crossover between a pair of semantic features $t_a^s$ and $t_b^s$. See detail in Algorithm \ref{AGSAlgo}. Finally, the semantic features $\tilde{t}^s$ derived from mutation and crossover are input into the proposed MKFNet to synthesize visual features $\tilde{x}^s$ and the cosine similarity $d$ is calculated using real visual features $x^s$ and synthesized visual features. The calculation process is formalized as:
\begin{equation}
   d=\frac{\tilde{x}^s\cdot x^s}{|\tilde{x}^s| |x^s|}
\end{equation}
Where $\cdot$ represents the dot product of vectors and $|x^s|$ represents the modulus of the vector $x^s$. The semantic features $\tilde{t}^s$ whose similarity are higher than the stability coefficient $\kappa_1$ ($d>\kappa_1$) will be viewed as enhanced semantic features $\tilde{t}^s_{enhanced}$ and added to the dataset to which semantic features $t_a^s$ and $t_b^s$ belong. The dataset $\mathcal{D}^{\bullet}$ will be updated as:
\begin{equation}
   \mathcal{D}^{\bullet}=(x_i^s, y_i^s, \{t_i^{\bullet}\bigcup \tilde{t}^s_{enhanced}\}), i=1,...,N^s
\end{equation}
Where $\tilde{t}^s_{enhanced}$ represents the high-stability semantic features generated by NFG.
% $\bullet$ is consistent with the labels of semantic features $t_a^s$ and $t_b^s$.
The semantic feature $\tilde{t}^s$ whose similarity is lower than the stability coefficient $\kappa_2$ ($d<\kappa_2$) will be viewed as novel semantic features $\tilde{t}^s_{novel}$ and added to the novel dataset $\mathcal{V}$. The novel dataset is defined as:
\begin{equation}
   \mathcal{V}=\{\mathcal{V}\bigcup \tilde{t}^s_{novel}\}
\end{equation}
Where $\tilde{t}^s_{novel}$ represents all low-stability semantic features generated by NFG and the low-confidence semantic feature is considered to have deviated to the unseen category. For $\tilde{t}_{enhanced}^s$ added to the corresponding datasets ($\mathcal{D}^{Species}, \mathcal{D}^{Genus}$ or $\mathcal{D}^{Family}$), they are continuously trained as the semantic features of the corresponding class label, and their loss function is formalized as:
\begin{equation}
\label{LNORM}
\begin{split}
    L_{ER}=&-\mathbb{E}[D(G(\tilde{t}_{enhanced}^s,z))]\\
    &-\mathbb{E}[\Sigma_{n=1}^{N^s}y_k \log(D(G(\tilde{t}_{enhanced}^s,z)))]
\end{split}
\end{equation}
Where, $N^s$ represents the total number of seen classes. For $\tilde{t}_{novel}^s$ in the novel dataset, they have no visual features and category labels. We use unsupervised novel regularization $L_{NR}$ to optimize the generative model, so that the generator can learn more unseen information. Its regularization term $L_{NR}$ is defined as:

\begin{equation}
\label{LNR}
L_{NR}=\mathbb{E}[D(G(\tilde{t}_{novel}^s,z))]+\lambda L_2[D(G(\tilde{t}_{novel}^s,z)), \tilde{y}]
\end{equation}
Where the first term tricks the discriminator to classify synthesized visual features from generator as fake and the second term is $L_2$ regularization to calculate the distance from the class distribution to the uniform distribution $\tilde{y}$. The $\tilde{y}$ is defined as a vector whose dimension is the number of seen classes $K^s$ and all values are $\frac{1}{K^s}$. $\lambda$ is penalty coefficient.

\begin{algorithm}
    \caption{The processes of mutation and crossover.}
    \label{AGSAlgo}
    \KwIn{$t_a^s, t_b^s$, the dimension of semantic feature $d$}
    \KwOut{semantic features $v_{mut}$ and $v_{cross}$}

    Random sampling a mutation coefficient $r_1$ and a crossover coefficient $r_2$ from uniform distribution $U(0,1)$.

    Random selecting $d*r_1$ and $d*r_2$ locations $loc_1$ and $loc_2$ for mutation and crossover, respectively.

    Let the semantic feature of mutation and crossover be $v_{mut}=t_a^s$ and $v_{cross}=t_a^s$ respectively.

    % 变异
    \For{$i=1;i \le d*r_1;i++$}
    {
        Sampling a random number $r$ from $U(0,1)$

        if $v_{mut}[loc_1[i]]\neq 0$: $v_{mut}[loc_1[i]] *= r$

        else: $v_{mut}[loc_1[i]] += r$
    }
    % 交叉
    \For{$i=1;i \le d*r_2;i++$}
    {
        $v_{cross}[loc_2[i]] = t_b^s[loc_2[i]]$
    }
\end{algorithm}

\subsection{Learning Scheme}
\label{TrainTest}
After we obtain the three datasets of different knowledge $\mathcal{D}^{s,Species}$, $\mathcal{D}^{s,Genus}$ and $\mathcal{D}^{s,Family}$, we select three sample points $(x^{s, \bullet}, y^{s, \bullet}, t^{s, \bullet})$ from them and input their semantic features into corresponding generator to obtain the synthesized visual features $\tilde{x}^{Species}, \tilde{x}^{Genus}$ and $\tilde{x}^{Family}$. Note that the three selected sample points must belong to the same Family. Then, the knowledge regularization $L_{KR}^{\bullet}$ of each generator is calculated according to the synthesized visual features. Next, these synthesized visual features are all propagated forward to the fusion model to get the final visual features $\tilde{x^s}$. Finally, the synthesized visual features $\tilde{x^s}$ are input into the discriminator and the loss function $L_{FM}$ of the fusion model, the loss function $L_D$ of the discriminator, and the loss function $L_G^{\bullet}$ of the generator are calculated. After iterative training of the above steps $N_{NFG}$ times, NFG with two regularization terms $L_{ER}$ and $L_{NR}$ is used to enrich semantic information and learn the information deviated to unseen classes. The training procedure of the proposed MKFNet is showed in Algorithm \ref{training}.

In testing, the visual features in unseen classes can be synthesized by three generators conditioned on a given unseen semantic feature $t^u$, as $\tilde{x}^{u, Family}=G_{Family}(t^u, z), \tilde{x}^{u, Genus}=G_{Genus}(t^u, z)$ and $\tilde{x}^{u, Species}=G_{Species}(t^u, z)$. And the final visual features $\tilde{x}$ can be obtained by adaptive fusion module. By this way, we can generate a large of synthesized visual features by sampling different $z$ for the sample text $t^u$. Finally, the class label of unseen visual feature $x^u$ can be determined by finding the label corresponding to the synthesized visual feature that is the most similar to the real visual feature.

\begin{algorithm}[!htb]
    \caption{The training procedure of the proposed MKFNet.}
    \label{training}
    \KwIn{the maximal loops $Step$, the start loop of NFG $N_{NFG}$, the stability coefficient $\kappa_1$, $\kappa_2$.}

    % Divide original dataset $D$ to three datasets $D^{s, Species}$, $D^{s, Genus}$ and $D^{s, Family}$ using Eq. \ref{DS}.
    Obtain $D^{s, Species}$, $D^{s, Genus}$ and $D^{s, Family}$ using Eq. \ref{DS}.

    Initialize the novel dataset $\mathcal{V}=\emptyset$.

    Calculate visual centers $\hat{x}^{Species}, \hat{x}^{Genus}, \hat{x}^{Family}$ for each class in $D^{s, Species}$, $D^{s, Genus}$ and $D^{s, Family}$, respectively.

      \For{$i=1;i \le Step;i++$}
      {
      \If{i $>N_{NFG}$}
      {
        % Execute NFG according to Algorithm \ref{AGSAlgo} and obtain new semantic features $\tilde{t}$.
        obtain new semantic features $\tilde{t}$ using Algorithm \ref{AGSAlgo}

        Get the visual features $\tilde{x}$ with $\tilde{t}$ by using Eq. \ref{fusionFormla}.

        Calculate the cosine similarity $d$ between $\tilde{x}$ and $x$.

        \If {$d>\kappa_1$} {
            Add $\tilde{t}$ to $\mathcal{D}^{s,Species}$, $\mathcal{D}^{s,Genus}$, or $\mathcal{D}^{s,Family}$.
        }

        \If {$d<\kappa_2$} {
            Add $\tilde{t}$ to $\mathcal{V}$.
        }
      }
      \For{$j=1;j \le 5;j++$}
      {
      Sample a minibatch of images $x$, matching texts $t^{s,\bullet}$, random noise $z$.

      % $\tilde{x}^{Species}, \tilde{x}^{Genus}, \tilde{x}^{Family} \leftarrow G_{Species}(t^{s,S},z), G_{Genus}(t^{s,G},z), G_{Family}(t^{s,F},z) $.
      $\tilde{x}^{\bullet}\leftarrow G_{\bullet}(t^{s,\bullet},z)$.

      Get the final visual features $\tilde{x}$ by using Eq. \ref{fusionFormla}.

      Calculate the discriminator loss $L_D$ by $x$ and $\tilde{x}$.

      Update discriminator parameters using Adam.
      }

      Sample a mini-batch of class labels $y$, matching texts $t^{s,\bullet}$, random noise $z$.

      Get the final visual features $\tilde{x}$ by using Eq.\ref{fusionFormla}.

      Calculate $L_{KR}^{\bullet}$ by Eq. \ref{LCRS}, \ref{LCRG}, \ref{LCRF} and $L_G^{\bullet}$ by Eq. \ref{GLossSpecies}.

      Calculate $L_{ER}$ by using Eq. \ref{LNORM}.

      Sample texts $\tilde{t}_{novel}$ from $\mathcal{V}$.

      Generate final visual features $\tilde{x}$ using $\tilde{t}_{novel}$ and calculate $L_{NR}$ according to Eq. \ref{LNR}.

      Calculate loss function $L_{FM}$ by using Eq. \ref{LFM}.

      Update parameters of generators using Adam($\bigtriangledown L_G^{\bullet}$).

      Update parameters of adaptive fusion module using Adam($\bigtriangledown L_{FM}$)

      %\For{$k=1;k \le N^s;k++$}
      %{
      %  $L_{tr}^{species}=\frac{1}{N^s}\|G(t_k, z)-\tilde{x}^{spec}_{y_k}\|^2$

      %  $L_{tr}^{genus}=\frac{1}{N^s}\|G(t_k, z)-\tilde{x}^{gen}_{y_k}\|^2$

       % $L_{tr}^{family}=\frac{1}{N^s}\|G(t_k, z)-\tilde{x}^{fam}_{y_k}\|^2$
      %}

      %$L_{tr}=\lambda_s L_{tr}^{species}+\lambda_g L_{tr}^{genus}+\lambda_f L_{tr}^{family}$

      %Compute the generator loss $L_G$ (which add taxonomy regularization $L_{tr}$) using Eq. 9.

      %Update the generator parameters using Adam($\bigtriangledown L_G$).

      }
\end{algorithm}

% 4. 实验
\section{Experiments}
% The generative ZSL methods, Baseline17 (ACGAN) \cite{ShlensConditional} and Baseline18 (GAZSL) \cite{Zhu_2018_CVPR}, are selected as the baseline in the experiment.

\subsection{Experimental Settings}
\label{ExperimentalSettings}
\textbf{Datasets:} We use seven common datasets to evaluate the proposed approach, which are three text-based recognition datasets and four attribute-based recognition datasets, respectively. The former include Caltech-UCSD-Birds 200-2011 (\textbf{CUB}) \cite{cub_dataset}, North America Birds (\textbf{NAB}) \cite{Horn2015Building} and Oxford 102 Flowers (\textbf{FLO}) \cite{nilsback2008automated}; The latter use GBU-setting \cite{xian2018zero} and include \textbf{CUB-Att} \cite{cub_dataset}, Animals with Attributes 1 (\textbf{AwA1}) \cite{lampert2009learning}, Animals with Attributes 2 (\textbf{AwA2}) \cite{Xian2017ZeroShotLC}, attributes Pascal and Yahoo (\textbf{aPY}) \cite{Farhadi2009Describing}.
For CUB and NAB datasets, two schemes \cite{elhoseiny2017link} are proposed to split the classes into training/testing (in total four benchmarks): Super-Category-Shared (SCS) or easy split and Super Category-Exclusive Splitting (SCE) or hard split. These two splits represent the similarity between the seen and unseen classes, so the former represents a higher similarity than the latter.
For CUB-Att, AwA1, AwA2 and aPY datasets, the standard zero-shot splits provided by \cite{Xian2017ZeroShotLC} is adopted. For FLO dataset, the splits provided by \cite{nilsback2008automated} is used.
The statistical information about these datasets is given in Table.\ref{dataset_info}.

\begin{table}
\centering
\caption{The statistical information of nine benchmarks. $|\mathcal{T}|$ represent the dimension of semantic feature. att? equals Y to represent semantic feature based on attribute, otherwise N for semantic feature based on Wikipedia text. $|\mathcal{X}|$ represent the number of images. $|\mathcal{Y}|$ represent the number of all classes, $|\mathcal{Y}^s|$ represent the number of classes in training+validation, and $|\mathcal{Y}^u|$ represent the number of test classes.}
\begin{tabular}{p{1.8cm}|p{0.65cm}p{0.4cm}p{0.65cm}p{0.4cm}p{1.cm}p{0.5cm}}
\toprule[1pt]
%下面是使用的数据集概况
\textbf{Dataset} & $|\mathcal{T}|$ & att? & $|\mathcal{X}|$ & $|\mathcal{Y}|$ & $|\mathcal{Y}^s|$ & $|\mathcal{Y}^u|$ \\ \midrule[1pt]
CUB(SCS) &7551  &N  &11788  &200  &120+30  &50  \\ % easy
CUB(SCE) &7551  &N  &11788  &200  &130+30  &40  \\ % hard
NAB(SCS) &13217  &N  &48562  &404  &200+123  &81  \\ % easy
NAB(SCE) &13217  &N  &48562  &404  &200+123  &81  \\ % hard
FLO &1024  &N  &8189  &102  &62+20  &20  \\
CUB-Att &312  &Y  &11788  &200  &100+50  &50  \\
AwA1 &85  &Y  &30475  &50  &27+13  &10  \\
AwA2 &85  &Y  &37322  &50  &27+13  &10  \\
aPY &64  &Y  &15339  &32  &15+5  &12   \\
% SUN &102  &Y  &37322  &645  &580+65  &72  \\
\bottomrule[1pt]
\end{tabular}
\label{dataset_info}
\end{table}

\textbf{Semantic Representation:} We use the raw Wikipedia articles extended by \cite{elhoseiny2017link} for both CUB and NAB. Like the textual representation method of \cite{Zhu_2018_CVPR,elhoseiny2019creativity}, we use the same preprocessing method and use Term Frequency-Inverse Document Frequency (TF-IDF) vectors as the semantic representation of Wikipedia articles. For FLO, we use the fine-grained visual descriptions collected by \cite{reed2016learning}. The 1024-dim character-based CNN-RNN \cite{reed2016learning} features are extracted from fine-grained visual descriptions (10 sentences per image). None of the $\mathcal{Y}^u$ sentences are seen during training the CNN-RNN. We build per-class sentences by averaging the CNN-RNN features that belong to the same class. Refer to \cite{xian2018feature,yu2020episode} for more details. For CUB-Att, AwA1, AwA2, and aPY datasets, we directly obtain the category attributes provided in the original dataset as semantic labels.

\textbf{Visual Representation:} For CUB and NAB, fine-grained visual features are extracted by using the part-based FC layer in VPDE-net \cite{zhang2017learning}. See \cite{Zhu_2018_CVPR,elhoseiny2019creativity,xian2018feature,Xian2017ZeroShotLC} for details. There are seven image-parts be detected in CUB dataset, which are head, back, belly, breast, leg, wing and tail. In NAB dataset, the "leg" part is missing because there is no annotation for this part in the original dataset. At last, the full dimensions of visual features in CUB and NAB are 3584D (512D $\times$ 7 parts) and 3072D (512D $\times$ 6 parts) respectively. Following \cite{Zhu_2018_CVPR,elhoseiny2019creativity,xian2018feature,Xian2017ZeroShotLC}, we use the top pooling units of the ResNet-101 \cite{he2016deep} pre-trained on ImageNet-1K as the visual features for the rest of datasets. Thus, those visual features can be represented as a 2048D vectors. For FLO, CUB-Att, AwA1, AwA2 and aPY, the 2,048-dimensional visual features are directly extracted by ResNet-101 \cite{he2016deep} pretrained on ImageNet.

%\textbf{Concept Division with Taxonomy:}
% The SUN dataset, which covers a large variety of environmental scenes, places and the objects within, has no taxonomy information. We generalize the Family-Genus-Species hierarchy of the taxonomy to the three-layers scene hierarchy provided by SUN dataset from https://vision.princeton.edu/projects/2010/SUN/hierarchy/.
%For datasets used in experiments, the taxonomy information is searched in https://encyclopedia.thefreedictionary.com/. We can obtain the Species by class name corresponding to category label. Then the corresponding Family and Genus can also be obtained through the Family-Genus-Species hierarchy. At last, all samples in datasets are labeled by Family, Genus and Species.

\textbf{Evaluation Metrics:} We use three metrics \cite{Chao2016An, Zhu_2018_CVPR, elhoseiny2019creativity, min2020domain, han2020learning} widely used in evaluating ZSL recognition performance: Top-1 unseen class accuracy, Area under Seen-Unseen curve (AUSUC) and harmonic mean ($H$). The first one is standard ZSL evaluation, which evaluates the performance on single domain (unseen domain). And the last two are generalized ZSL evaluation, which evaluates the performance on two domains (seen and unseen domains). The harmonic mean is denoted by $H=(2S\times U)/(S+U)$. $S$ and $U$ are the Top-1 Accuracies for seen and unseen domains.

\textbf{Implementation Details:}
We implement our model with neural networks using PyTorch. For each generator ($G_{Family}$, $G_{Genus}$ and $G_{Species}$) and discriminator $D$ in MKFNet, the skeleton and parameter settings are consistent with GAZSL \cite{Zhu_2018_CVPR} which will be used as the baseline of our model for comparison. We use Adam solver with $\beta_1=0.5$ and $\beta_2=0.9$. We empirically set the total number of iterations $N_{NFG}=3$, the stability coefficient $\kappa_1=0.8, \kappa_2=0.2$. The full source code of our model can be found at github \footnote{https://github.com/a494456818/MKFNet-NFG}.
% and period of initiating genetic strategy $n_{period}=2000$.
% Code is available at https://www.github.com/.

\subsection{Comparison with state-of-the-art methods}
\textbf{Experimental Results on CUB and NAB: }
\hspace{0.5em} The CUB and NAB datasets are used to evaluate the performance of the proposed method on Wikipedia articles. Table \ref{wiki_zsl_top1_acc} shows the state-of-the-art results on CUB and NAB with easy and hard splittings. We observe that the proposed method achieves significant improvements over the Baseline18 and the state-of-the-art methods in terms of both Top-1 accuracy and Seen-Unseen AUC. Specially, in the hard splitting of CUB and NAB, the proposed method achieves 54.4\% and 40.7\% improvements on Top-1 accuracy metric, 65.5\% and 46.6\% improvements on Seen-Unseen AUC metric, compared with Baseline18. In the easy splitting of CUB and NAB, the proposed method achieves 8.7\% and 3.1\% improvements on Top-1 accuracy metric, 14.4\% and 13.7\% improvements on Seen-Unseen AUC metric, compared with Baseline18.
Compared with the state-of-the-art methods, the proposed method achieves 3.7\% and 10.4\% improvements on Top-1 accuracy metric, 0.7\% and 15.2\% improvements on Seen-Unseen AUC metric respectively in easy and hard splittings of CUB. In easy and hard splittings of CUB, the proposed method achieves an improvement of up to 30.1\% under Top-1 and AUC metrics.
% Compared with the state-of-the-art methods, the consistency of the proposed method is improved under Top-1 and AUC metrics across CUB and NAB datasets, up to 30.1\%.

% 对比表
\begin{table*}[htp]
	\centering
	\caption{Result of ZSL Image Classification on \textbf{CUB} and \textbf{NAB}. We reproduce the results of GDAN, whose source code is available online. The red value represent the increasing number compared with the Baseline18.}
	\begin{tabular}{p{4.5cm}|p{1.1cm}p{1.1cm}p{1.1cm}p{1.1cm}|p{1.1cm}p{1.1cm}p{1.1cm}p{1.1cm}}
		\toprule[1pt]
		&  \multicolumn{4}{c|}{Top-1 Accuracy}       &  \multicolumn{4}{c}{Seen-Unseen AUC}     \\
		&  \multicolumn{2}{c}{CUB} & \multicolumn{2}{c|}{NAB} &  \multicolumn{2}{c}{CUB} & \multicolumn{2}{c}{NAB}     \\
		methods & Easy & Hard & Easy & Hard & Easy & Hard & Easy & Hard \\ \midrule[1pt]
		
		% Baseline17 \cite{ShlensConditional} &15.7  &6.6  &12.1  &4.1  &3.1  &3.0  &2.1  &0.2  \\

		WAC-Linear\cite{Elhoseiny2013Write} &27.0  &5.5  &-  &-  &23.9  &4.9  &23.5  &-  \\
		WAC-Kernel\cite{ElhoseinyWrite} &33.5  &7.7  &11.4  &6.0  &14.7  &4.4  &9.3  &2.3  \\
		ESZSL\cite{Romera2015An} &28.5  &7.4  &24.3  &6.3  &18.5  &4.5  &9.2  &2.9  \\
		ZSLNS\cite{QiaoLess} &29.1  &7.3  &24.5  &6.8  &14.7  &4.4  &9.3  &2.3  \\
		SynC$_{fast}$\cite{Changpinyo2016Synthesized} &28.0  &8.6  &18.4  &3.8  &13.1  &4.0  &2.7  &3.5  \\
		ZSLPP\cite{ElhoseinyLink} &37.2  &9.7  &30.3  &8.1  &30.4  &6.1  &12.6  &3.5  \\
        % GAZSL \cite{Zhu_2018_CVPR} &43.7  &10.3  &35.6  &8.6  &35.4  &8.7  &20.4  &5.8  \\
        FeatGen\cite{xian2018feature} &43.9  &9.8  &36.2  &8.7  &34.1  &7.4  &21.3  &5.6  \\
		CIZSL\cite{elhoseiny2019creativity} &44.6  &14.4  &36.6  &9.3  &39.2  &11.9  &24.5  &6.4  \\
        CANZSL\cite{chen2020canzsl} &45.8  &14.3  &38.1  &8.9  &40.2  &12.5  &\textbf{25.6}  &6.8  \\
	%	Baseline'18-2$+$CIZSL\cite{2019arXiv190401109E} &44.2  &12.1  &36.3  &9.8  &37.4  &9.8  &24.7  &6.2  \\
        GDAN\cite{Huang_2019_CVPR} &44.2  &13.7  &\textbf{38.3}  &8.7  &38.7  &10.9  &24.1  &5.9  \\
        %MLSE\cite{DingL19} &-  &-  &-  &-  &-  &-  &-  &-  \\
        %VSE-S*\cite{zhu2019generalized} &-  &-  &-  &-  &-  &-  &-  &-  \\
        \midrule[1pt]
	    Baseline18 \cite{Zhu_2018_CVPR} &43.7  &10.3  &35.6  &8.6  &35.4  &8.7  &20.4  &5.8  \\
        % GAN-CST\cite{} &46.1  &14.1  &38.6  &10.4  &40.5  &12.7  &24.9  &7.8  \\
        CKL-TR\cite{xie2021cross} &45.8  &15.1  &36.8  &10.0  &40.2  &12.2  &22.3  &7.8  \\
        MKFNet-NFG &\textbf{47.5}$^{\textcolor{red}{+3.8}}$  &\textbf{15.9}$^{\textcolor{red}{+5.6}}$  &37.4$^{\textcolor{red}{+1.8}}$  &\textbf{12.1}$^{\textcolor{red}{+3.5}}$  &\textbf{40.5}$^{\textcolor{red}{+5.1}}$  &\textbf{14.4}$^{\textcolor{red}{+5.7}}$  &23.2$^{\textcolor{red}{+2.8}}$  &\textbf{8.5}$^{\textcolor{red}{+2.7}}$  \\

		% \textbf{Baseline18+CKL} &43.9$^{\textcolor{red}{+0.2}}$  &11.5$^{\textcolor{red}{+1.2}}$  &35.7$^{\textcolor{red}{+0.1}}$  &9.0$^{\textcolor{red}{+0.4}}$  &38.6$^{\textcolor{red}{+3.2}}$  &10.7$^{\textcolor{red}{+2.0}}$  &22.7$^{\textcolor{red}{+2.3}}$  &5.7$^{\textcolor{blue}{-0.1}}$ \\
		% \textbf{Baseline18+CKL+TR} &45.8$^{\textcolor{red}{+2.1}}$  &\textbf{15.1}$^{\textcolor{red}{+4.8}}$  &36.8$^{\textcolor{red}{+1.2}}$  &10.0$^{\textcolor{red}{+1.4}}$  &40.2$^{\textcolor{red}{+4.8}}$  &12.2$^{\textcolor{red}{+3.5}}$  &22.3$^{\textcolor{red}{+1.9}}$  &7.8$^{\textcolor{red}{+2.0}}$  \\
		\midrule[1pt]
	\end{tabular}
	\label{wiki_zsl_top1_acc}
\end{table*}

\textbf{Experimental Results on CUB-Att, AwA1, AwA2, FLO and aPY: }In these five benchmarks, we used new metrics ($U$, $S$ and $H$) to evaluate the performance of our approach. In order to evaluate that our approach is still effective under different semantic representations, CUB-Att, AwA1, AwA2 and aPY with GBU-setting is applied to change the semantic representation from Wikipedia articles to attributes. The evaluation results of $S$, $U$ and $H$ on CUB-Att, AwA1, AwA2, FLO and aPY are given in Table \ref{GBU_results}. We observe that the proposed approach surpassed state-of-the-art methods on most evaluation metrics in these five datasets. Specially in $H$ metric, the $H$ value improvement on AwA1 increases from 72.1\% to 76.4\%, on AwA2 increases from 67.1\% to 74.0\%, on FLO from 76.0\% to 83.7\%. Compared with the generative ZSL methods \cite{Zhu_2018_CVPR,xian2018feature,xian2019f,Huang_2019_CVPR,li2019leveraging}, the performance of our method has been consistently improved on almost all evaluation metrics, and the highest improvement is 23.4\% in the $U$ metric of the FLO dataset. Remarkably, compared with Baseline18, the proposed method achieves a very significant improvement, e.g., 172.6\% and 112.7\% improvements of $U$ metric on FLO and aPY. This demonstrates the effectiveness of the proposed method in generating more discriminative visual features.

% 对比表
\begin{table*}[htp]
	\centering
	\caption{Results of comparison with state-of-the-arts methods in Generalized ZSL (GZSL) scenario on four classification benchmarks. $U$ and $S$ are the Top-1 accuracies tested on unseen classes and seen classes, respectively, in GZSL. H is the harmonic mean of U and S.}
	\begin{tabular}{p{2.8cm}|p{0.5cm}p{0.5cm}p{0.5cm}|p{0.5cm}p{0.5cm}p{0.5cm}|p{0.5cm}p{0.5cm}p{0.5cm}|p{0.5cm}p{0.5cm}p{0.5cm}|p{0.5cm}p{0.5cm}p{0.5cm}}
		\toprule[1pt]
		&  \multicolumn{3}{c|}{CUB-Att} & \multicolumn{3}{c|}{AwA1} & \multicolumn{3}{c|}{AwA2} & \multicolumn{3}{c|}{FLO} &  \multicolumn{3}{c}{aPY} \\
		methods & $U$ & $S$ & $H$ & $U$ & $S$ & $H$ & $U$ & $S$ & $H$ & $U$ & $S$ & $H$ & $U$ & $S$ & $H$ \\ \midrule[1pt]
		
		Baseline18 \cite{Zhu_2018_CVPR} & 31.7 & 61.3 & 41.8 & 29.6 & 84.2 & 43.8 & 35.4 & 86.9 & 50.3 & 28.1 & 77.4 & 41.2 & 14.2 & 78.6 & 24.0 \\

        CLSWGAN \cite{xian2018feature} & 43.7 & 57.7 & 49.7 & 57.9 & 61.4 & 59.6 & 56.1 & 65.5 & 60.4 & 59.0 & 73.9 & 65.6 & - & - & -  \\
        SE-ZSL \cite{kumar2018generalized} & 41.5 & 53.3 & 46.7 & 56.3 & 67.8 & 61.5 & 58.3 & 68.1 & 62.8 & - & - & - & - & - & -  \\
        LisGAN \cite{li2019leveraging} & 46.5 & 57.9 & 51.6 & 52.6 & 76.3 & 62.3 & 47.0 & 77.6 & 58.5 & 57.7 & 83.8 & 68.3 & - & - & -  \\
        f-VAEGAN-D2 \cite{xian2019f} & 48.4 & 60.1 & 53.6 & 57.6 & 70.6 & 63.5 & - & - & - & 56.8 & 74.9 & 64.6 & - & - & -  \\
        CADA-VAE \cite{schonfeld2019generalized} & 51.6 & 53.5 & 52.4 & 57.3 & 72.8 & 64.1 & 55.8 & 75.0 & 63.9 & - & - & - & 31.5 & 57.1 & 40.6  \\
        RELATION NET \cite{sung2018learning} & 38.1 & 61.1 & 47.0 & 31.4 & 91.3 & 46.7 & 30.0 & \textbf{93.4} & 45.3 & 50.8 & 88.5 & 64.5 & - & - & -   \\
		GDAN \cite{Huang_2019_CVPR} & 39.3 & 66.7 & 49.5 & - & - & - & 32.1 & 67.5 & 43.5 & - & - & - & 30.4 & 75.0 & 43.4   \\
        DAZLE \cite{huynh2020fine} & 56.7 & 59.6 & 58.1 & - & - & - & 60.3 & 75.0 & 67.1 & - & - & - & - & - & -   \\
        RFF-GZSL \cite{han2020learning} & 50.6 & 79.1 & 61.7 & 59.5 & 91.6 & 72.1 & - & - & - & 61.3 & 88.8 & 72.5 & - & - & -   \\
        OCD \cite{keshari2020generalized} & 53.2 & 60.2 & 56.5 & - & - & - & 59.5 & 73.4 & 65.7 & - & - & - & - & - & -   \\
        DVBE \cite{min2020domain} & 53.2 & 60.2 & 56.5 & - & - & - & 63.6 & 70.8 & 67.0 & - & - & - & 32.6 & 58.3 & 41.8   \\
        ASPN \cite{lu2020attentive} & 50.7 & 61.5 & 55.6 & 58.0 & 85.7 & 69.2 & 46.2 & 87.0 & 60.4 & 67.3 & 87.4 & 76.0 & - & - & -   \\
        APNet \cite{liu2020attribute} & 55.9 & 48.1 & 51.7 & \textbf{76.6} & 59.7 & 67.1 & 54.8 & 83.9 & 66.4 & - & - & - & \textbf{32.7} & 74.7 & 45.5   \\
        \midrule[1pt]
        CKL-TR \cite{xie2021cross} & 57.8 & 50.2 & 53.7 & 61.4 & \textbf{93.2} & 74.0 & 61.2 & 92.6 & 73.7 & - & - & - & 30.8 & 78.9 & 44.3   \\
        MKFNet-NFG & \textbf{58.9} & \textbf{64.8} & \textbf{61.7} & 65.0 & 92.9 & \textbf{76.4} & \textbf{61.9} & 92.2 & \textbf{74.0} & \textbf{76.6} & \textbf{92.4} & \textbf{83.7} & 30.2 & \textbf{94.0} & \textbf{45.7}   \\
		\midrule[1pt]
	\end{tabular}
	\label{GBU_results}
\end{table*}

\subsection{Ablation Studies}
In this section, we performed ablation experiments on several important components of the proposed method. To evaluate the impact of the adaptive fusion module, we use simple features summing operation instead of the adaptive fusion module, denoted as MKFNet-base. To evaluate the effectiveness of the MKFNet, we use CKL-TR \cite{xie2021cross} (which is our previous work) as the baseline because our method is improved based on it and uses the new proposed generative model. The terms MKFNet and MKFNet-NFG indicate using NFG method and not using NFG method respectively and they all uses the adaptive fusion module. The results are reported in Table.\ref{ablationStudy}. Compared with MKFNet-base, our MKFNet has achieved a consistent improvement in all metric, which represents that the adaptive fusion module can produce more appropriate weights effectively and automatically to fuse better visual features. we can see that MKFNet-base has a higher accuracy in the seen classes, 67.1\% than 63.3\%, indicating that the visual features generated by the simple features summing has poor generalization because it has a little over-fitting in the seen class. Compared with CKL-TR, our MKFNet has achieved a consistent improvement in H metric, which shows that the proposed generative model can synthesize more generalized visual features. Especially in CUB-Att, the H metric is improved from 53.7\% to 60.0\%. A reasonable reason is that the fine-grained CUB dataset is more densely distributed in the visual space, and the fusion of visual features of higher knowledge level is beneficial to alleviate this phenomenon and expand the class spacing. In addition, the base version MKFNet-base of MKFNet outperforming CKL-TR in some metrics can also illustrate the effectiveness of the proposed model. For example, the H metric is improved from 53.7\% to 58.7\% and 74.0\% to 75.2\% on CUB-Att and AwA1 datasets, respectively. Although the H metric on the aPY dataset has decreased slightly, from 44.3\% to 42.8\%, MKFNet-base retains a high accuracy in seen classes, increasing from 78.9\% to 90.6\%. Furthermore, compared with MKFNet, MKFNet-NFG achieved consistent improvements, with H metric ranging from 60.0\% to 61.7\%, 75.4\% to 76.4\%, and 44.9\% to 45.7\% on the CUB-Att, AwA1 and aPY datasets, which shows that the NFG method does enable the model to learn some new discriminative features.
In general, the overall trend in Table.\ref{ablationStudy} can be observed that with the combination of each component, the performance has been consistently improved. This obviously shows the effectiveness of each component.
% In the table, the horizontal axis represents the proposed method with different components, and the vertical axis represents the evaluation metric, which is Top-1 accuracy or AUSUC. The overall trend is that with the combination of each component, the performance has been consistently improved. This obviously shows the effectiveness of our various components.

% 消融实验的表格
\begin{table}[htp]
\newcommand{\tabincell}[2]{\begin{tabular}{@{}#1@{}}#2\end{tabular}}  %导言区
	\centering
	\caption{The influence of different components on GZSL results. MKFNet-base is the base version of MKFNet, which uses simple summing operation instead of the adaptive fusion module of MKFNet. MKFNet and MKFNet-NFG represent using NFG method and not using NFG method, respectively.}
    \begin{tabular}{p{1.68cm}|p{0.35cm}p{0.35cm}p{0.4cm}|p{0.35cm}p{0.35cm}p{0.35cm}|p{0.35cm}p{0.35cm}p{0.35cm}}
		\toprule[1pt]
		&  \multicolumn{3}{c|}{CUB-Att}  &  \multicolumn{3}{c|}{AwA1}  &  \multicolumn{3}{c}{aPY}     \\
		methods & U & S & H & U & S & H & U & S & H \\ \midrule[1pt]
		% Baseline17 \cite{ShlensConditional} &15.7  &6.6  &12.1  &4.1  &3.1  &3.0  &2.1  &0.2  \\
        CKL-TR &57.8  &50.2  &53.7  &61.4  &\textbf{93.2}  &74.0  &\textbf{30.8}  &78.9  &44.3  \\
        MKFNet-base &52.2  &\textbf{67.1}  &58.7  &63.3  &92.5  &75.2  &28.0  &90.6  &42.8  \\
        MKFNet   &56.8  &63.6  &60.0  &63.6  &92.6  &75.4  &29.8  &90.7  &44.9  \\
        MKFNet-NFG   &\textbf{58.9}  &64.8  &\textbf{61.7}  &\textbf{65.0}  &92.9  &\textbf{76.4}  &30.2  &\textbf{94.0}  &\textbf{45.7}  \\
		\midrule[1pt]
	\end{tabular}
	\label{ablationStudy}
\end{table}

\subsection{Visualization Results}

\begin{figure}
\centering
\includegraphics[width=8.5cm,height=4cm]{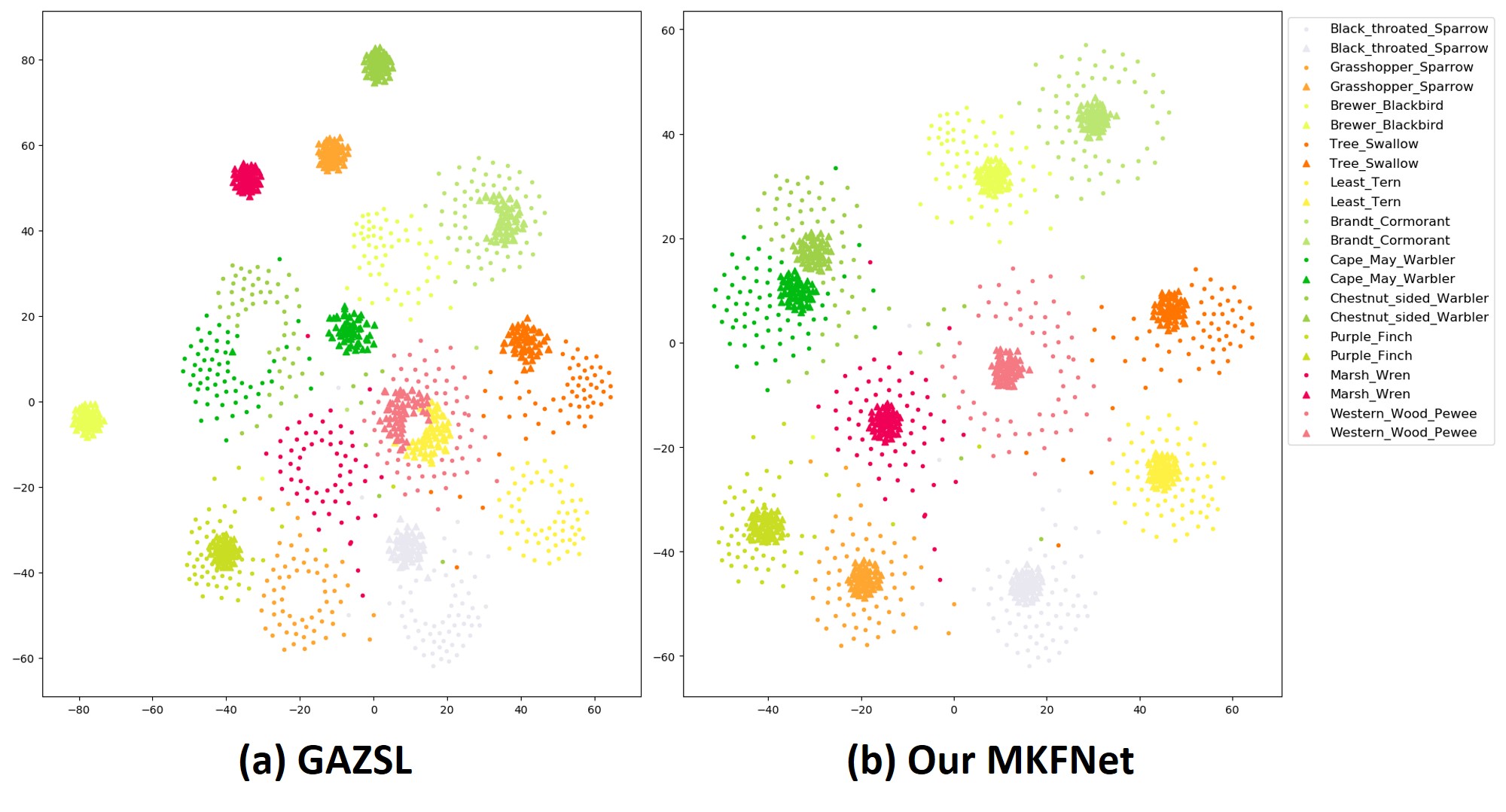}
\caption {The t-SNE visualization of real and synthesized visual features. The circle and triangle represents the real visual features and the synthesized visual features, respectively. The different colors represent different classes. }
\label{t-SNE}
\end{figure}

\textbf{Feature Visualization: }To investigate the difference between the real visual features and the synthesized visual features of the unseen classes, we visualize the real visual features and synthesized visual features by using t-SNE method and compare the visualization result of the proposed approach with GAZSL \cite{Zhu_2018_CVPR} on CUB dataset. Specially, for each unseen class on CUB, we use our method and GAZSL to synthesize 60 visual features. The visualization results of GAZSL and ours are shown in Fig.\ref{t-SNE}(a) and Fig.\ref{t-SNE}(b), respectively. As shown in Fig.\ref{t-SNE}(a), there are serious differences in the visual feature distribution between real samples and synthesized samples, especially Marsh Wren, Grasshopper Sparrow and Cape May Warbler et al. In Fig.\ref{t-SNE}(b), it is obviously that the synthesized visual features of our method follow the distribution of the real visual features. This proves that the proposed method can learn more general knowledge and deviate to the unseen class.

\begin{figure}
\centering
\includegraphics[width=8cm,height=4cm]{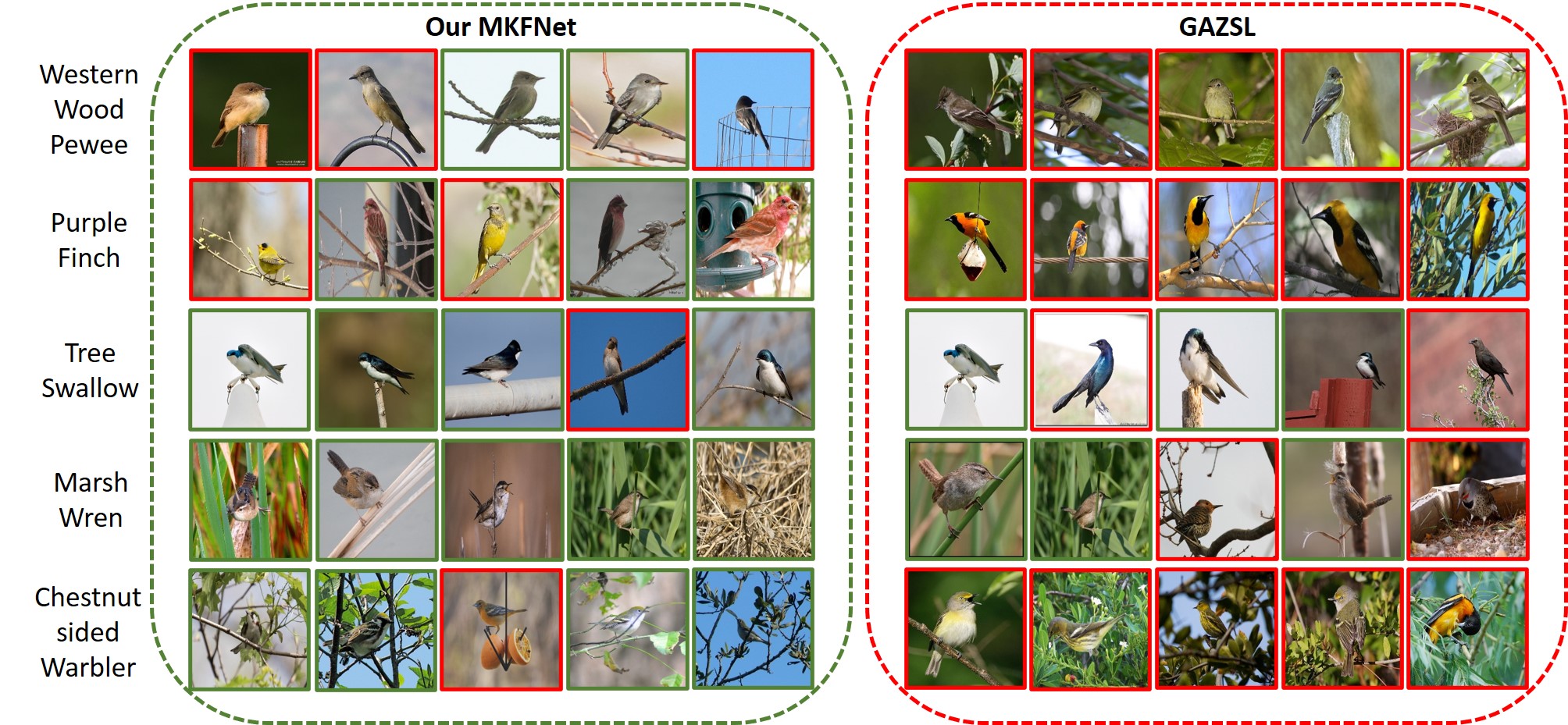}
\caption {The ZSL retrieval results of the proposed method in CUB dataset. Each row contains the Top-5 retrieved images of a specific class. The image with a green box and a red box represent correct and wrong retrieval, respectively.}
\label{retrievalResult}
\end{figure}

\textbf{Result of ZSL Image Retrieval: }We compare the proposed approach with GAZSL \cite{Zhu_2018_CVPR} on the image retrieval task on CUB. For each semantic feature in CUB, we generate 60 visual features and calculate the visual center point. According to these visual central points, the cosine similarity is calculated with the true visual features, and the Top-5 samples with the highest similarity are selected as the retrieved images. The retrieval results are shown in Fig.\ref{retrievalResult}. The green box and red box represent the correct retrieval image and the wrong retrieval image, respectively. The retrieval results show that the proposed approach is more accurate that GAZSL, demonstrating that our approach can learn more discriminative feature and generate more appropriate synthesized visual features.

\subsection{Discuss}
In the previous experiments, we used a large number of datasets to indirectly prove that the proposed method achieved very promising performance. To further discuss why the proposed method is effective, we conducted two additional experiments.
% 在前面的实验中，我们使用了大量的数据集间接地证明了提出的方法取得了非常有希望的性能。为了进一步讨论提出的方法为什么有效，我们额外进行了2个实验。

\begin{figure}[!htp]
\centering
\subfigure[FLO dataset]{
    \includegraphics[width=0.145\textwidth]{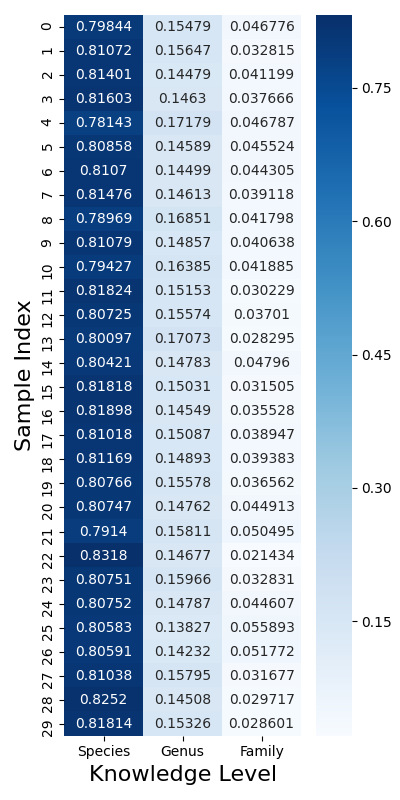}
}
\subfigure[AwA1 dataset]{
    \includegraphics[width=0.145\textwidth]{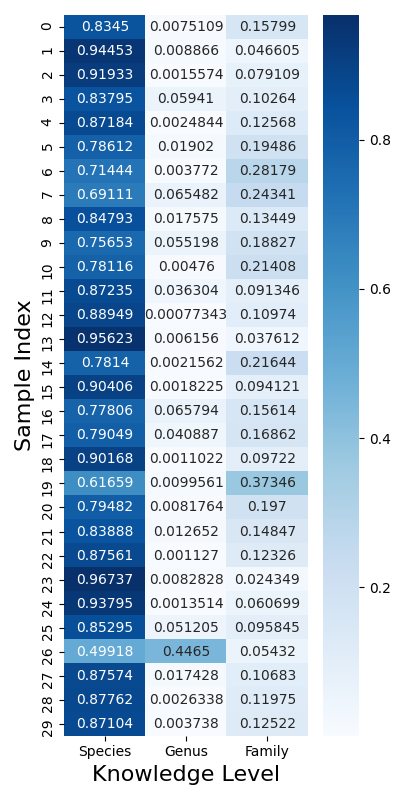}
}
\subfigure[aPY dataset]{
    \includegraphics[width=0.145\textwidth]{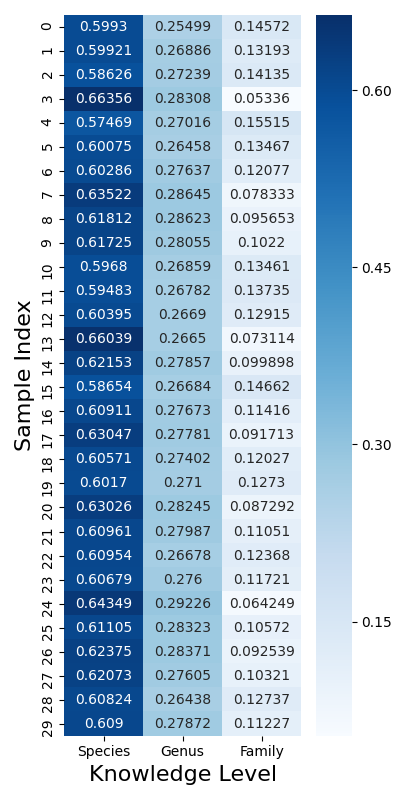}
}
\caption{The importance weight of each knowledge in different benchmarks.}
\label{discuss_weight}
\end{figure}

\textbf{Effectiveness of Adaptive Fusion Module in MKFNet.} In this work, we proposed Multi-Knowledge Fusion Network (MKFNet) with adaptive fusion module to meet the domain-shift challenge. The idea of MKFNet is to fuse multiple semantic features from different knowledge. Different knowledge have different universality and specificity. Their fusion can make the generated samples have stronger generalization ability. In order to explore the effectiveness of each knowledge in different benchmarks, we visualize the importance weights of different knowledge domains, as shown in Fig.\ref{discuss_weight}. Each subgraph in Fig.\ref{discuss_weight} shows the importance weight of FLO, AWA1 and APY datasets in different knowledge domains. In order to display more information, we select 30 samples from unseen classes randomly and input them into MKFNet, and record the importance weight of different knowledge in each sample. The horizontal axis of each subgraph represents different knowledge (Species, Genus, and Family), and the vertical axis represents the importance weights of 30 samples. Obviously, we can observe that in different datasets, the importance weights of different knowledge domains are all greater than zero. This shows that different knowledge domains have played a role, and further proves the effectiveness of the adaptive fusion module in MKFNet. By introducing higher-level knowledge, we can generate more generalized samples to alleviate the domain-shift problem.
%不同的知识，他们具有不同的普适性和特异性，的融合可以使生成的样本具有更强的泛化能力。为了探索每个知识在不同benchmark发挥的作用，我们可视化不同知识域的重要性权重，如图2所示。在Fig.\ref{discuss_weight}中的每个子图分别展示了FLO，AWA1和APY数据集在不同知识领中的重要性权重。为了显示更多信息，我们从随机从未见类中筛选30个样本输入到MKFNet中，并记录不同知识在每个样本中的重要性权重。每个子图的横轴表示不同的知识，纵轴表示30个样本的重要性权重。显然，我们能够观察到在不同的数据集中，不同知识领域的重要性权重均大于0。这显示了不同知识领域都发挥了作用和在MKFNet中自适应融合模型的有效性。

\begin{figure}[!htp]
\centering
\subfigure[AwA1 dataset]{
    \includegraphics[width=0.225\textwidth]{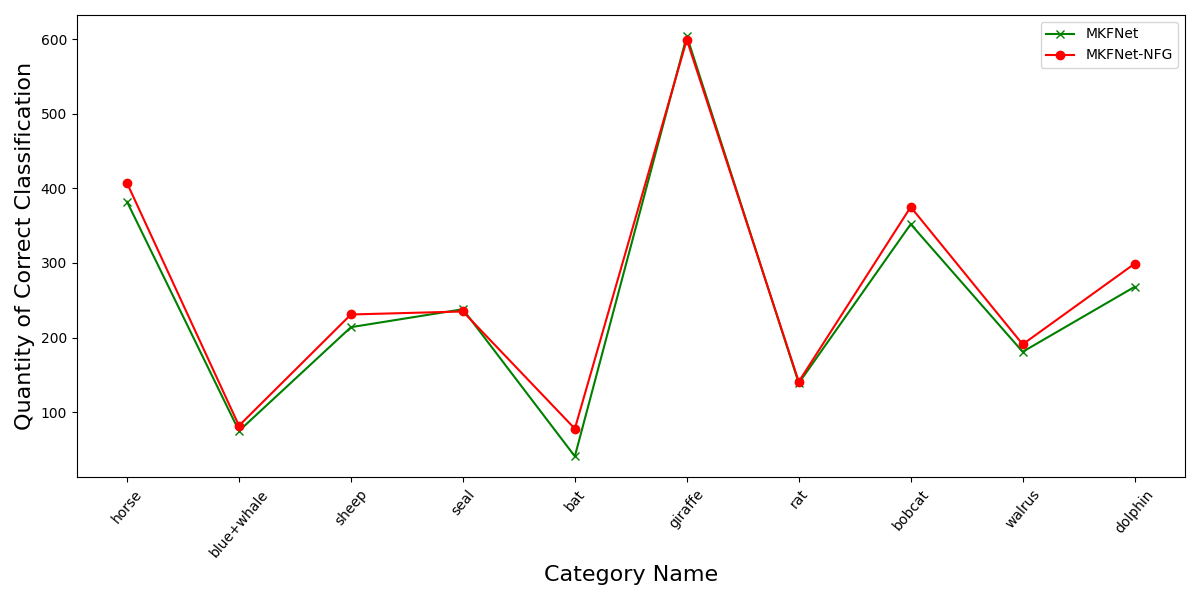}
}
\subfigure[AWA2 dataset]{
    \includegraphics[width=0.225\textwidth]{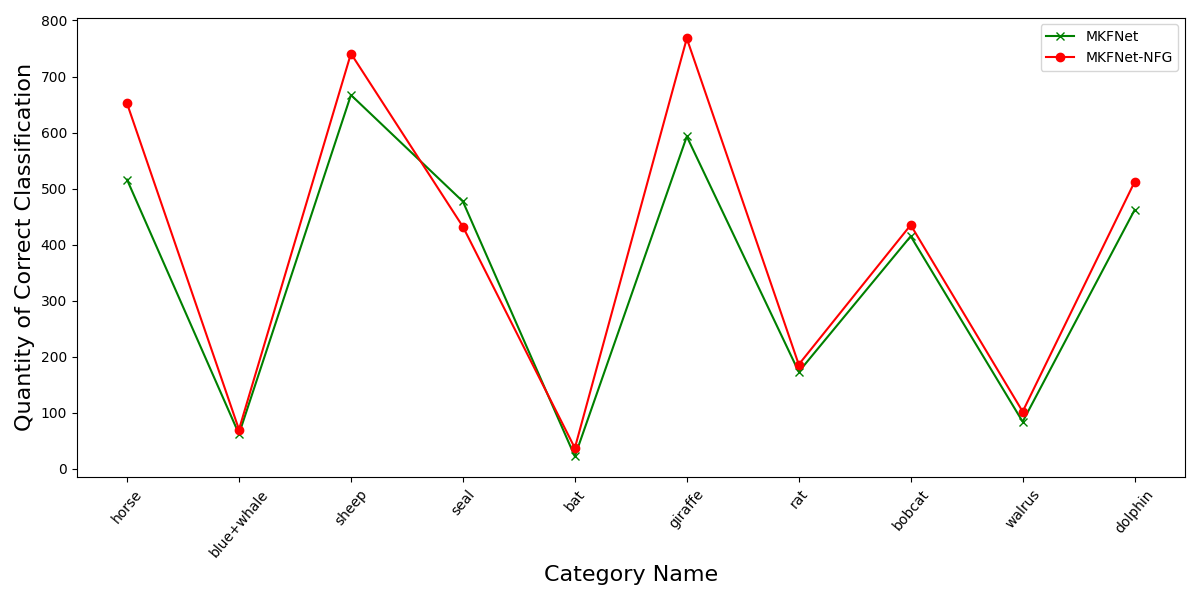}
}
\subfigure[CUB-Att dataset]{
    \includegraphics[width=0.225\textwidth]{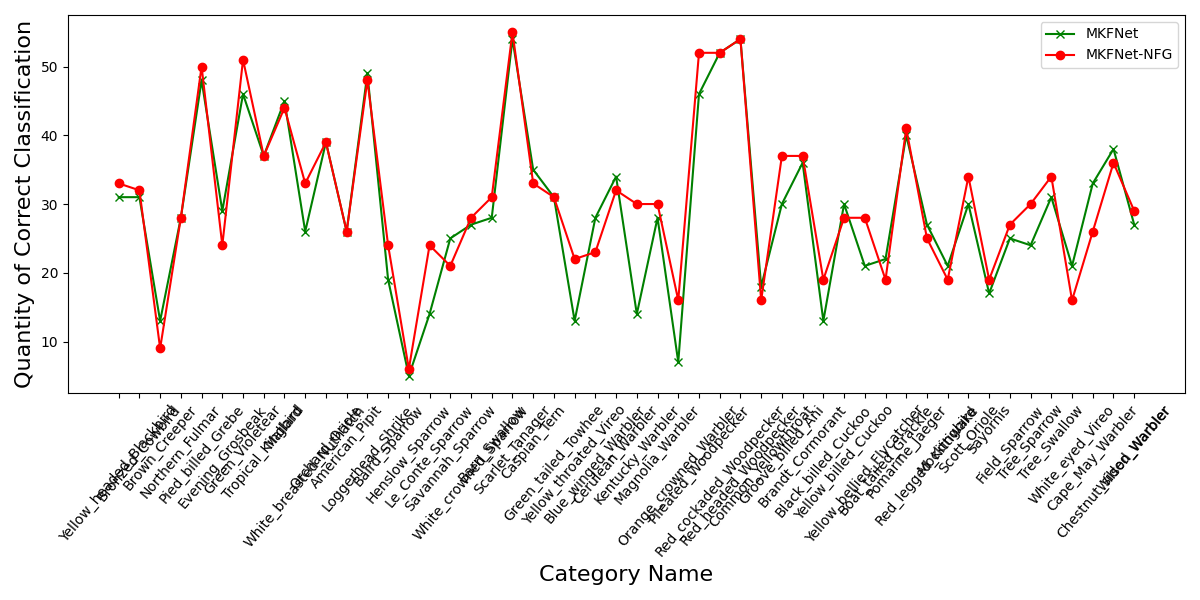}
}
\subfigure[FLO dataset]{
    \includegraphics[width=0.225\textwidth]{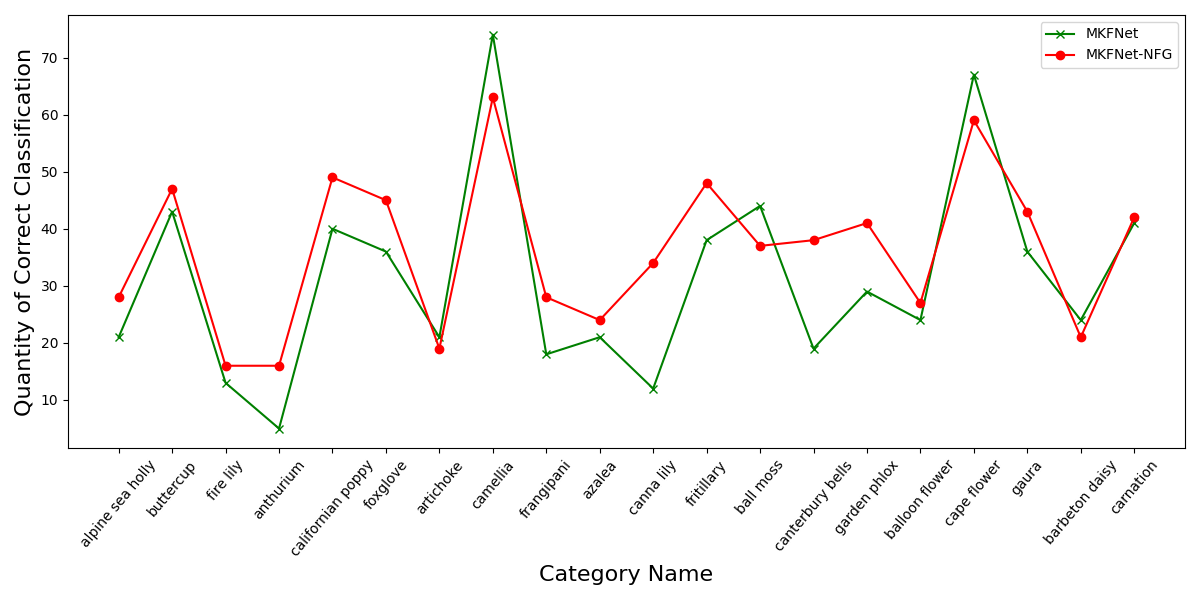}
}
\caption{The influence of using NFG method (MKFNet-NFG) and not using NFG method (MKFNet) on the number of unseen classes correctly identified. }
\label{discuss2_num}
\end{figure}

\textbf{Effectiveness of New Feature Generator in MKFNet.} In order to further prove the positive impact of the New Feature Generator (NFG) on the classification of each unseen category, we counted the number of correct classifications for each category on AwA1, AwA2, CUB-Att and FLO datasets. As shown in each sub-graph in the Fig.\ref{discuss2_num}, the x-axis represents the name of the unseen category, and the y-axis represents the number of samples correctly classified as the unseen category. An obvious trend is to use NFG's MKFNet to be able to identify more unseen samples. Especially for the AwA1 and AwA2 datasets, MKFNet-NFG has a higher recognition rate than MKFNet on almost every unseen category. This phenomenon shows that NFG can generate new discriminative features for model learning to synthesize generalized visual features.
% 为了进一步证明NFG方法对每个未见类分类的正面影响，我们统计了每个类别分类正确的数量对于AwA1, AwA2, CUB-Att and FLO datasets。如图\ref{discuss2_num}中的每个子图所示，x轴表示未见类的名称，y轴表示正确分类未见类的样本数量。一个明显的趋势是使用NFG的MKFNet能够识别更多的未见类样本。特别是对于AwA2和CUB-Att数据集，使用NFG的MKFNet几乎在每一个未见类上的识别率都高于MKFNet。这一现象说明NFG能够产生新的具有判别性的特征供模型学习。
% 挑选出因为GA算法识别出来的样本

% 5. 结论
\section{Conclusion}
In this paper, we proposed a novel generative zero-shot learning approach to address the problems of semantic insufficiency and domain-shift problems in ZSL, where Multi-Knowledge Fusion Network (MKFNet) and $L_{ER}$ in New Feature Generator (NFG) are proposed to enhance semantic features by using cross-concept knowledge, and $L_{NR}$ in AGS and $L_{KR}$ are proposed improves the intersection of synthesized visual features and unseen visual features.
The experiment results demonstrate that our approach is superior to baselines and several state-of-the-art methods.

% 致谢
\section*{Acknowledgement}
This work was supported in part by the Natural Science Foundation of China (NSFC) under Grant 61876166 and Grant 61663046, in part by the Yunnan Applied Fundamental Research Project under Grant 2016FB104, in part by the Yunnan Provincial Young Academic and Technical Leaders Reserve Talents under Grant 2017HB005, in part by the Program for Yunnan High Level Overseas Talent Recruitment, and in part by the Program for Excellent Young Talents of Yunnan University.

% 附录
%\appendices
%\section{}

\bibliography{ref}
\bibliographystyle{IEEEtran}

\end{document}